\documentclass[final,3p,times,twocolumn]{elsarticle}

\usepackage{amssymb}

\usepackage{lineno}

\usepackage{amsmath}
\usepackage{times}
\usepackage{epsfig}
\usepackage{graphicx}

\usepackage{supertabular}
\usepackage{rotating}
\usepackage{caption}
\usepackage{subcaption}

\journal{Applied Energy}

\begin{document}

\begin{frontmatter}


\cortext[cor1]{E-mail address: qp208@cam.ac.uk (Quentin Paletta)}

\title{ECLIPSE : Envisioning CLoud Induced Perturbations in Solar Energy}

\author[label1,label2]{Quentin Paletta\corref{cor1}}
\author[label1]{Anthony Hu}
\author[label2]{Guillaume Arbod}
\author[label1]{Joan Lasenby}

\address[label1]{Department of Engineering, University of Cambridge, UK}
\address[label2]{ENGIE Lab CRIGEN, France}




\begin{abstract}
Efficient integration of solar energy into the electricity mix depends on a reliable anticipation of its intermittency. A promising approach to forecast the temporal variability of solar irradiance resulting from the cloud cover dynamics is based on the analysis of sequences of ground-taken sky images or satellite observations. Despite encouraging results, a recurrent limitation of existing deep learning approaches lies in the ubiquitous tendency of reacting to past observations rather than actively anticipating future events. This leads to a frequent temporal lag and limited ability to predict sudden events. To address this challenge, we introduce ECLIPSE, a spatio-temporal neural network architecture that models cloud motion from sky images to not only predict future irradiance levels and associated uncertainties, but also segmented images, which provide richer information on the local irradiance map. We show that ECLIPSE anticipates critical events and reduces temporal delay while generating visually realistic futures. The model characteristics and properties are investigated with an ablation study and a comparative study on the benefits and different ways to integrate auxiliary data into the modelling. The model predictions are also interpreted through an analysis of the principal spatio-temporal components learned during network training.
\end{abstract}


\end{frontmatter}

\section{Introduction}

The current energy transition towards an increased utilisation of renewable energy sources entails multiple challenges. Unlike conventional fossil fuels, the production of electricity from wind and the sun suffers from an inherent and substantial input variability, constituting a major limitation to their large scale integration to the energy mix. A key solution to increasing their usability lies in a better forecast of the incoming energy supply. By modelling the atmospheric perturbations responsible for this variability (wind, clouds, aerosols, temperature, precipitation, etc.), the future power fluctuations can be anticipated, and the response of the electricity system adapted accordingly~\cite{kumariLongShortTerm2021}. Applications benefiting from energy supply forecasting~\cite{inmanSolarForecastingMethods2013a, dasForecastingPhotovoltaicPower2018} range from smart grids~\cite{suchBatteryEnergyStorage2012} and energy storage management~\cite{padronAnalysisPumpedStorage2011}, to hybrid power plants~\cite{jamelAdvancesIntegrationSolar2013} or frequency control~\cite{khalidOptimalOperationWind2012}.
\vspace{0.5\baselineskip}

The flow of clouds over solar facilities constitutes the predominant source of variability in solar energy. For instance, a thick cloud obstructing the sun from view might, within a minute, largely eliminate the solar flux directly originating from the sun (as opposed to the residual diffuse irradiance), thus considerably reducing the local energy generation. However, the modelling of the cloud cover is a challenging task due to its dynamic nature, sporadic spatial distribution in the sky, all guided by complex natural phenomena. For these reasons, statistical methods based on historical data often fail to accurately predict such sudden events. In contrast, more promising approaches based on computer vision techniques, are better suited to recognise objects (sun, cloud, etc.) and thus determine corresponding solar conditions~\cite{marquezForecastingGlobalHorizontal2012a, berneckerContinuousShorttermIrradiance2014a, wangNeuralNetworkBased2018}.
\vspace{0.5\baselineskip}

\begin{figure}
\centering
\begin{minipage}[b]{0.21\textwidth}
    \includegraphics[width=1.1\textwidth]{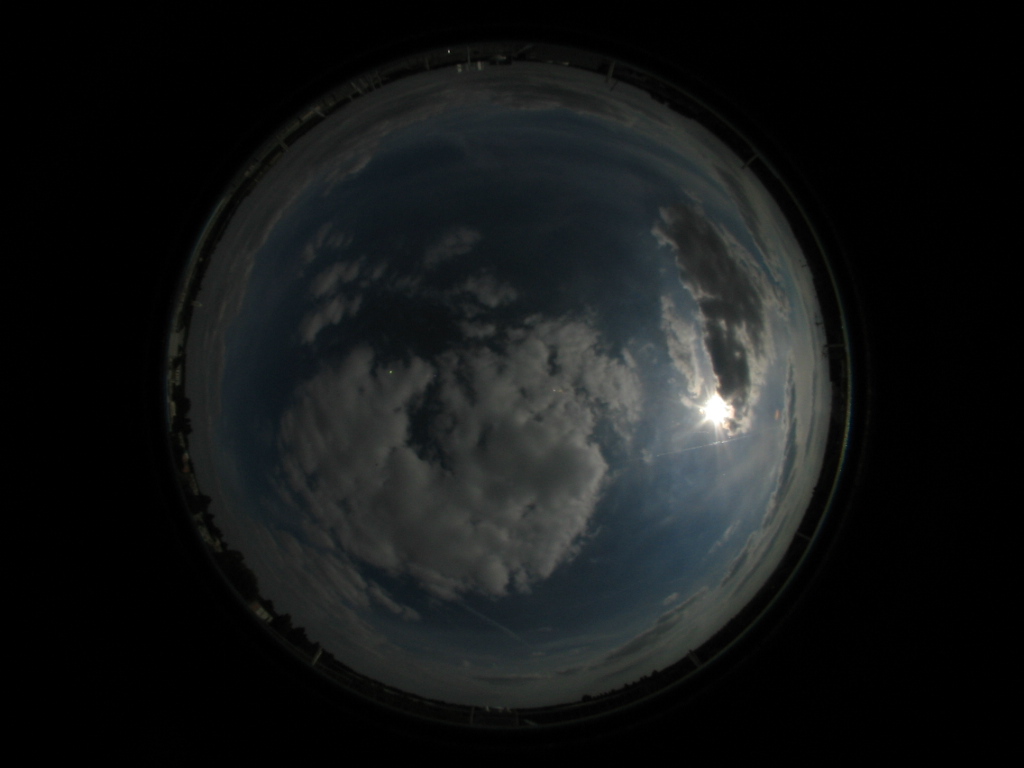}
    \label{fig:img_sh_2018_7_3_13_0}
  \end{minipage} 
  \begin{minipage}[b]{0.21\textwidth}
    \includegraphics[width=1.1\textwidth]{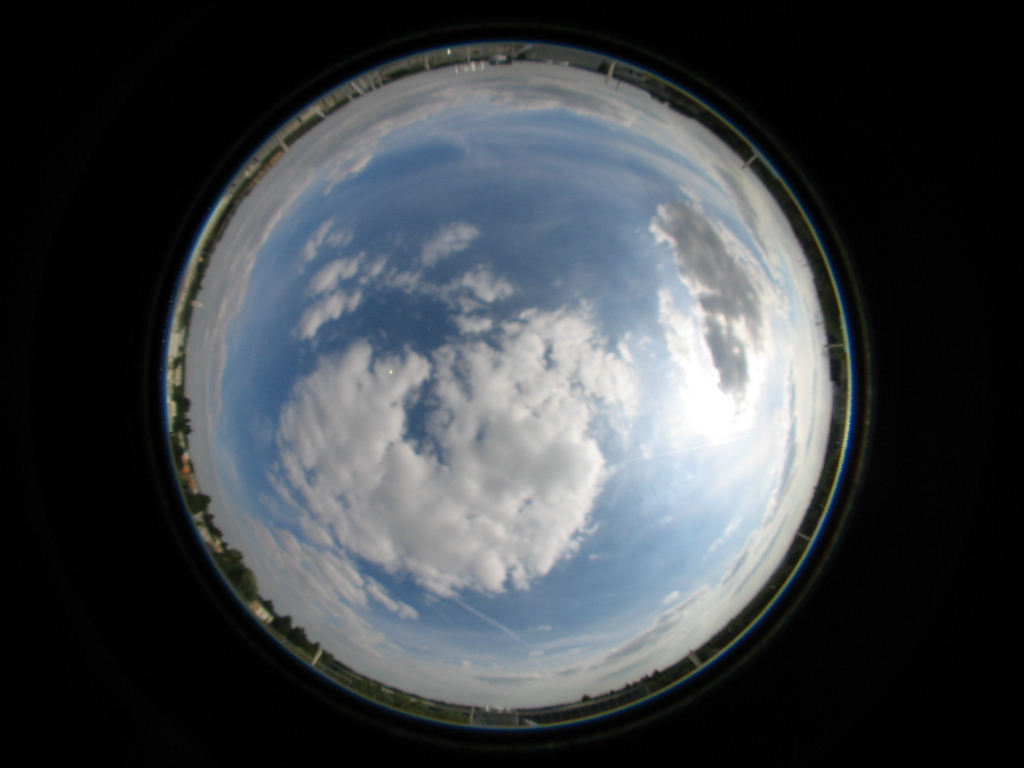}
    \label{fig:img_sun_204_2018_7_3_13_0}
  \end{minipage}
\vspace{-0.8\baselineskip}
\caption{Example of sky images taken by a fish-eye camera with short (left panel) and long (right panel) exposures (shared by SIRTA laboratory~\cite{sirta}).}
\label{fig:sky_images}
\end{figure}

Satellite imagery provides the best observations of the cloud cover for longer-term (up to several hours) computer vision based predictions~\cite{hamillShortTermCloudForecast1993, kurzrockReviewUseGeostationary2018, crosReliabilityPredictorsSolar2020}. This approach is able to  anticipate the displacements of larger clouds and consequently predict the future solar resource over a large area. However, its ability to anticipate high local irradiance changes caused by specific clouds is limited due to its low pixel resolution of around $1 \text{km}^2$ covering twice the surface of a typical solar farm. In contrast, for very short-term forecasting (up to 20-min), recent studies have focused on ground-taken sky images, which can provide a higher spatio-temporal resolution (10-sec to 2min, covering an area of up to 1.5 $\text{km}^2$ centered on the camera~\cite{kongHybridApproachesBased2020}) and are thereby better suited for accurate local predictions~\cite{yangSolarIrradianceForecasting2014}. More recently, machine learning approaches were developed to model the cloud cover dynamics from satellite observations~\cite{perezDeepLearningModel2021b, siPhotovoltaicPowerForecast2021a, nielsenIrradianceNetSpatiotemporalDeep2021}.
\vspace{0.5\baselineskip}

\textit{In situ} sky cameras equipped with fish-eye lenses are able to capture the surrounding sky with a high spatial resolution (see Figure~\ref{fig:sky_images}). From a sequence of sky images, cloud tracking techniques provide a tool for the modelling of the cloud cover dynamics used to anticipate incoming clouds~\cite{chowIntrahourForecastingTotal2011, ruizCloudtrackingMethodologyIntrahour2014a}. This can be performed using block matching~\cite{huangCloudMotionEstimation2013} or optical flow~\cite{Wood-Bradley2012}. Furthermore, one can generalise this local prediction to a wider neighbouring area by estimating the cloud induced shadow map and its resulting irradiance map from cloud properties~\cite{nouriDeterminationCloudTransmittance2019}. This can be done using multiple camera observations to produce a 3D model of the cloud cover relative to the ground~\cite{peng3DCloudDetection2015, blancShorttermForecastingHigh2017, kuhnDeterminationOptimalCamera2019, rodriguez-benitezAssessmentNewSolar2021b}. In recent years, increased availability of open source datasets~\cite{pedroComprehensiveDatasetAccelerated2019} and virtual datasets~\cite{kurtzVirtualSkyImager2017} have fostered interest in data-driven approaches, which have been shown to outperform traditional methods in many computer vision tasks~\cite{weinzaepfelDeepFlowLargeDisplacement2013, shiConvolutionalLSTMNetwork2015, heMaskRCNN2020}. A detailed review of the field can be found in~\cite{linRecentAdvancesIntrahour2022}.

\begin{figure*}
\centering    
\includegraphics[width=1\textwidth]{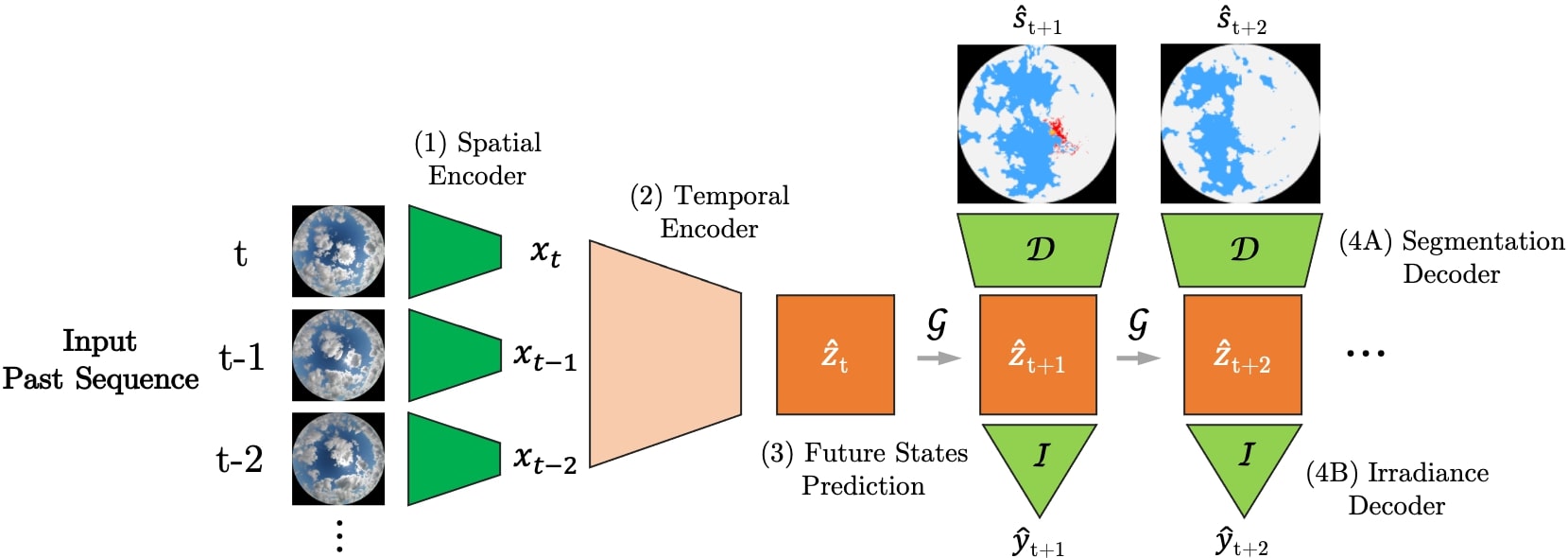}
\vspace{-0.5\baselineskip}
\caption{ECLIPSE is composed of five modules: (1) Spatial Encoder, (2) Temporal Encoder, (3) Future State Prediction module, (4A) Segmentation Decoder and (4B) Irradiance Decoder. It iteratively predicts future segmented images and corresponding irradiance levels \textit{I} from a compact representation of past sky images.}
\label{fig:model_architecture}
\end{figure*}

\vspace{1\baselineskip}
\paragraph{Contributions} We build on advances in models for video prediction \cite{wangPredRNNRecurrentNeural2017,lucPredictingDeeperFuture2017,clarkAdversarialVideoGeneration2019,huProbabilisticFuturePrediction2020} to propose a deep learning (DL) approach (Figure~\ref{fig:model_architecture}), learning cloud motion from sky images or satellite images and predicting both segmented images and their corresponding irradiance levels. We show that our model ECLIPSE - Envisioning CLoud Induced Perturbations in Solar Energy - better anticipates critical events by reducing prediction time lag~\cite{palettaBenchmarkingDeepLearning2021c}. Predicted state representations can be decoded into a visually realistic future, which provides rich operational information on the surrounding irradiance map, for instance, benefiting maintenance planning or future power output estimation. We extend the traditional binary (cloud/sky) segmentation approach with a new method based on an image-based sun tracker~\cite{palettaTemporallyConsistentImagebased2020} to determine sun masks in sky images, which can be exploited to improve the translation of a cloud map to the corresponding local irradiance map. This new approach addresses a key limitation of current deep learning methods as stated in the 2021 International Energy Agency Task 16 Report~\cite{senguptaBestPracticesHandbook2021a}: ``the difficulty of [...] interpreting the information [generated] to determine what is happening in the natural world'' by generating visual feedback on the modelled cloud cover dynamics (Section~\ref{section:future_state_predictions}), quantifying prediction uncertainties in specific weather conditions (Section~\ref{section:model_uncertainty}) and visually interpreting model predictions with a novel application of an explainable AI approach (Section~\ref{section:pca_spatiotemporal_features}).

\vspace{0.5\baselineskip}
Follow-up studies using ECLIPSE include a benchmark between different scene representations and data augmentation techniques~\cite{palettaSPIN2021} and a novel strategy to combine both sky and satellite observations in a single machine learning framework~\cite{palettaOmnivisionForecastingCombining2022}.

\vspace{-0.2\baselineskip}
\section{Related Work}

The recent application of neural networks to the modelling of solar energy variability has shown promising results~\cite{ahmedReviewEvaluationStateoftheart2020}. Artificial Neural Networks (ANNs) were first successfully trained to translate an image of the sky into its corresponding simultaneous irradiance value~\cite{wangNeuralNetworkBased2018, sunSolarPVOutput2018}. Convolutional Neural Network (CNN) models have been shown to recognise specific cloud patterns and adjust their prediction accordingly~\cite{wenDeepLearningBased2021, palettaConvolutionalNeuralNetworks2020}. Regarding the more challenging task of predicting future solar flux or solar energy production, numerous DL architectures have been shown to reach high quantitative performances relative to standard computer vision approaches: CNN~\cite{sunConvolutionalNeuralNetwork2018, fengSolarNetSkyImagebased2020a, fengConvolutionalNeuralNetworks2022}, CNN + Long short Term Memory (LSTM) networks~\cite{zhangDeepPhotovoltaicNowcasting2018, fengSolarNetSkyImagebased2020, palettaConvolutionalNeuralNetworks2020, wenDeepLearningBased2021, siddiquiDeepLearningApproach2019, palettaBenchmarkingDeepLearning2021c}, 3D-CNN~\cite{zhao3DCNNbasedFeatureExtraction2019, palettaBenchmarkingDeepLearning2021c}, implicit layers~\cite{leguenDeepPhysicalModel2020a}, Convolutional LSTM~\cite{kongHybridApproachesBased2020, palettaBenchmarkingDeepLearning2021c}.
\vspace{0.5\baselineskip}

Despite promising results based on common metrics such as the root Mean Square Error (RMSE), DL approaches applied to irradiance forecasting still face major limitations. In particular, traditional architectures suffer from a frequent forecast time lag, hence low anticipation skills~\cite{kongHybridApproachesBased2020, palettaBenchmarkingDeepLearning2021c}. Instead of actively anticipating future events (high irradiance shifts), neural networks tend to react to the current weather conditions in order to minimise the future risk of large errors. This translates into a heavy inertia and averaged forecasts. Resulting predictions seem to face a \textit{persistence barrier}, representing the inability to foresee events before they happen, i.e. to decrease time lag below the forecast horizon. In addition, contrary to traditional methods based on 3D cloud cover modelling~\cite{nouriEvaluationAllSky2020}, few DL studies predict future sky images, which are more valuable than single irradiance measurements in many industrial applications. Predicted RGB images by DL models tend to be fuzzy and unrealistic, especially in the circumsolar area, hence difficult to exploit to generate local irradiance maps~\cite{leguenDeepPhysicalModel2020}.

\vspace{0.5\baselineskip}
This paper is structured as follows: Section~\ref{section:methodology} details the methodology of the study. The main quantitative and qualitative results of our work are then presented in Section~\ref{section:results} and~\ref{section:eumetsat}. Finally, we conclude in Sections~\ref{section:conclusion} with an outlook on future research directions.
\vspace{0.5\baselineskip}

\section{Methodology}
\label{section:methodology}
\subsection{Model Architecture}

Contrarily to previous architectures trained end-to-end to predict a single irradiance value or sky image at a given time horizon~\cite{sunConvolutionalNeuralNetwork2018, zhangDeepPhotovoltaicNowcasting2018, zhao3DCNNbasedFeatureExtraction2019, siddiquiDeepLearningApproach2019, leguenDeepPhysicalModel2020a}, we recursively predict future states that are then regressed to future irradiance values (Figure~\ref{fig:model_architecture}). Predicting a sequence of future values instead of a single value allows our model to learn a representation that can detect rapid changes in solar flux, due to cloud occlusion for instance.
\vspace{0.5\baselineskip}

Moreover, these states are decoded into future sky semantic segmentation, which provides a much richer and stronger supervision signal compared to solely using the sparse irradiance signal. Segmented images can be translated to local irradiance maps, which are more informative for large solar farms than single location irradiance values. Compared to RGB images, segmented images are a more compact high-level representation abstracting the scene to focus on the most relevant information (the shape of a cloud but not its texture).

\vspace{0.5\baselineskip}
\textbf{Spatial Encoder} ($\text{Width} \times \text{Height} \times \text{Channels} : 128 \times 128 \times 3 \xrightarrow{} 16 \times 16 \times 128$) Spatial features are first extracted from RGB sky images through three spatial down-sampling convolutional layers: a 2D residual convolutional layer with 64 filters and kernel size 7 (2D convolutions, 2D batch normalisation, Relu activation and maxpooling), followed by two ResNet~\cite{heDeepResidualLearning2016} modules with 64 and 128 filters respectively. The spatial dimension of the $H\times W$ input images is downsampled by 8 (from $128 \times 128$ to $16 \times 16$) prior to the temporal encoder, through strides in the convolutions.

\vspace{0.5\baselineskip}
\textbf{Temporal Encoder ($16 \times 16 \times 128 \times t  \xrightarrow{} 16 \times 16 \times 110$)} The spatial features, denoted $(x_1, x_2, ..., x_t)$, are encoded into a spatio-temporal feature $z_t$ through Temporal Blocks~\cite{huProbabilisticFuturePrediction2020}. These blocks are made of separable 3D convolutional layers with both local (3D convolution) and global (3D pooling layers) modules. Global layers give spatio-temporal context to the learned representation $z_t$, and local layers model fine motion from the cloud cover dynamics. The feature $z_t$ has spatial dimension $(\frac{H}{8},\frac{W}{8})$ and 110 channels.

\vspace{0.5\baselineskip}
\textbf{Future State Prediction ($16 \times 16 \times 110 \xrightarrow{} H \times 16 \times 16 \times 110$)}
Through a recurrent convolutional module $\mathcal{G}$, we iteratively predict future states:

\vspace{-0.3\baselineskip}

\begin{equation}
    \hat{z}_{t+i+1} = \mathcal{G}(\hat{z}_{t+i})
\end{equation}

For $i=0,...,H-1$, with $H$ the horizon of future prediction and $\hat{z}_t = z_t$. This recurrent module $\mathcal{G}$ is a repetition of four blocks, each block containing a spatial Gated Recurrent Unit layer \cite{siamConvolutionalGatedRecurrent2017} followed by four 2D residual layers. The number of input and output channels, as well as the spatial dimensions are kept identical: $(110,\frac{H}{8},\frac{W}{8})$.
\vspace{0.5\baselineskip}

This flexible future state prediction strategy can be well accommodated to specific operation forecasting schemes imposed by system operators or transmission organisations in terms of lead time, time horizon and temporal resolution~\cite{fengConvolutionalNeuralNetworks2022}.

\vspace{0.5\baselineskip}
\textbf{Segmentation Decoder ($16 \times 16 \times 110 \xrightarrow{} 128 \times 128 \times 5$)} Each predicted future state is decoded into a future semantic segmentation of the sky $\hat{s}$ through an upsampling convolutional module $\mathcal{D}$:

\vspace{-0.3\baselineskip}
\begin{equation}
    \hat{s}_{t+i} = \mathcal{D}(\hat{z}_{t+i})
\end{equation}

$\mathcal{D}$ is a succession of three 2D residual layers with 256, 128 and 64 filters respectively, and three bilinear upsampling modules so that the output $\hat{s}_{t+i}$ has dimension $C\times H\times W$, with $C$ the number of semantic classes.

\vspace{0.5\baselineskip}
A detailed occlusion map of individual solar panels can be derived from segmented sky images. This provides richer information than a single use of local irradiance measurements for utility operations.

\vspace{0.5\baselineskip}
\textbf{Irradiance Module ($16 \times 16 \times 110 \xrightarrow{} 1$)} Additionally, each future state is mapped to an irradiance value $\hat{y}$ through the irradiance module $\mathcal{I}$. 
\vspace{-0.3\baselineskip}
\begin{equation}
    \hat{y}_{t+i} = \mathcal{I}(\hat{z}_{t+i})
\end{equation}

$\mathcal{I}$ contains two convolutional layers with 64 filters each, a 2D average pooling layer (that averages over all spatial dimensions), followed by two densely connected layers with 64 nodes and 1 node respectively. See \ref{section:architecture} for more technical details on the architecture.

\subsection{Dataset}
\label{section:dataset}

\begin{figure*}
\centering
\begin{minipage}[b]{0.19\textwidth}
    \includegraphics[width=0.985\textwidth]{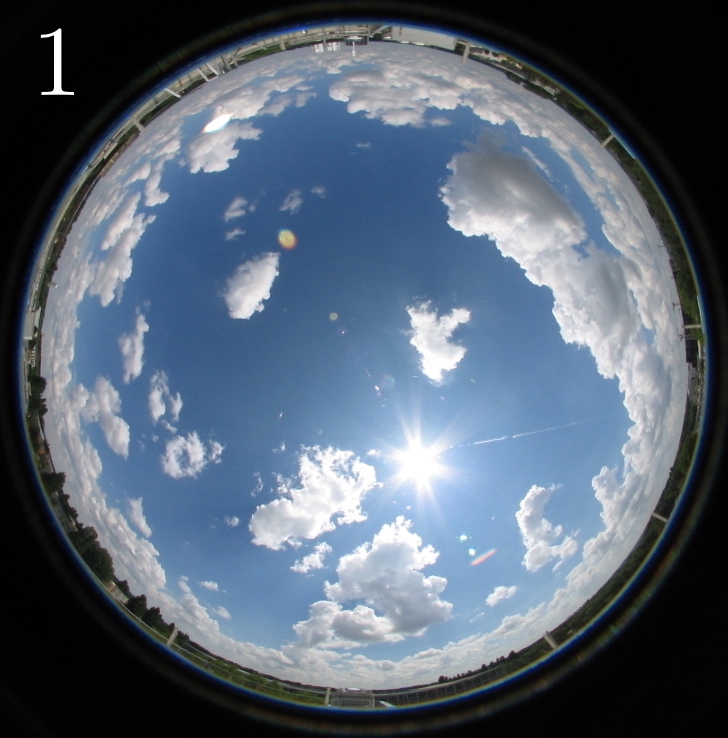}
    \label{fig:sky_image_raw}
  \end{minipage} 
\begin{minipage}[b]{0.19\textwidth}
    \includegraphics[width=1\textwidth]{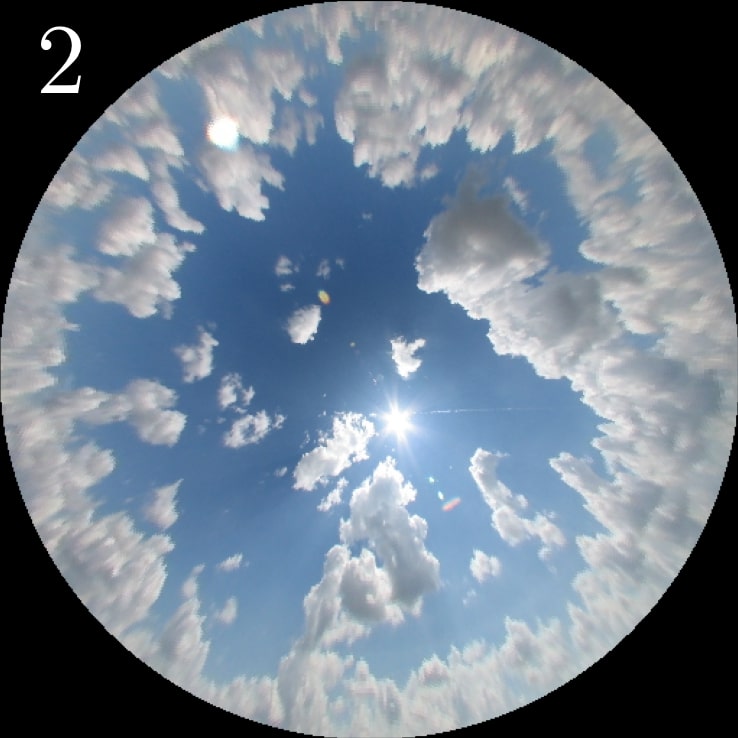}
    \label{fig:sky_image_undistorted}
  \end{minipage}
  \begin{minipage}[b]{0.19\textwidth}
    \includegraphics[width=1\textwidth]{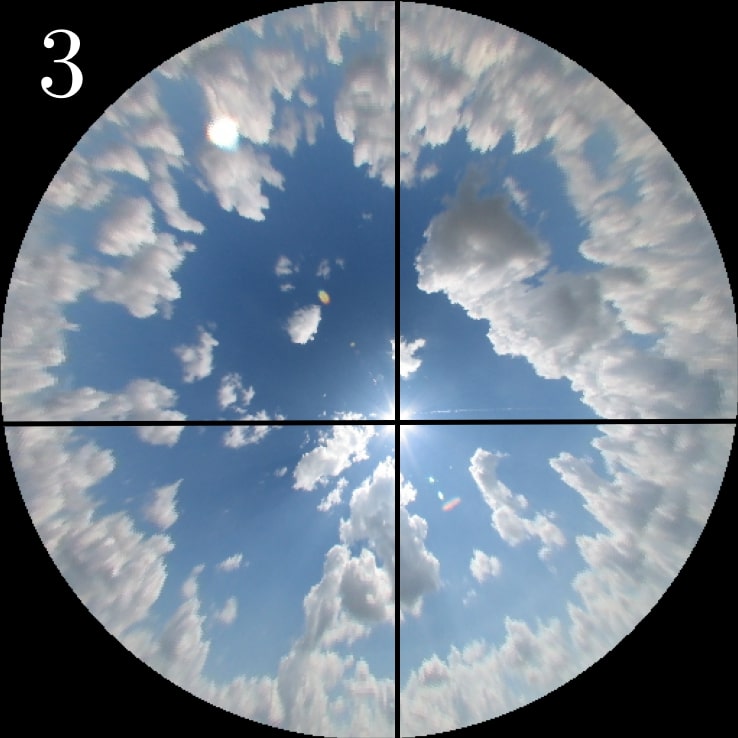}
    \label{fig:sky_image_target}
  \end{minipage}
   \begin{minipage}[b]{0.19\textwidth}
    \includegraphics[width=1\textwidth]{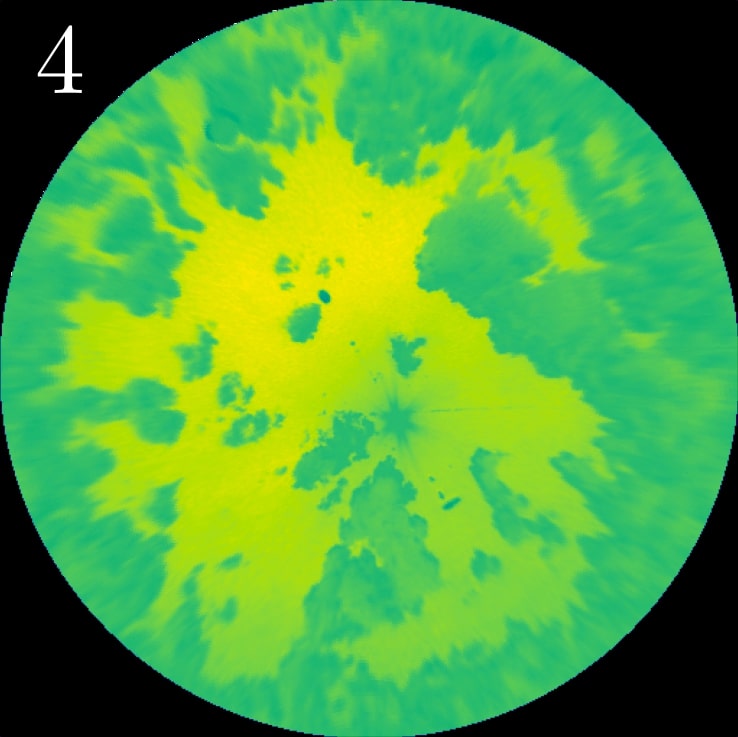}
    \label{fig:sky_image_nRBR}
  \end{minipage} 
  \begin{minipage}[b]{0.19\textwidth}
    \includegraphics[width=1\textwidth]{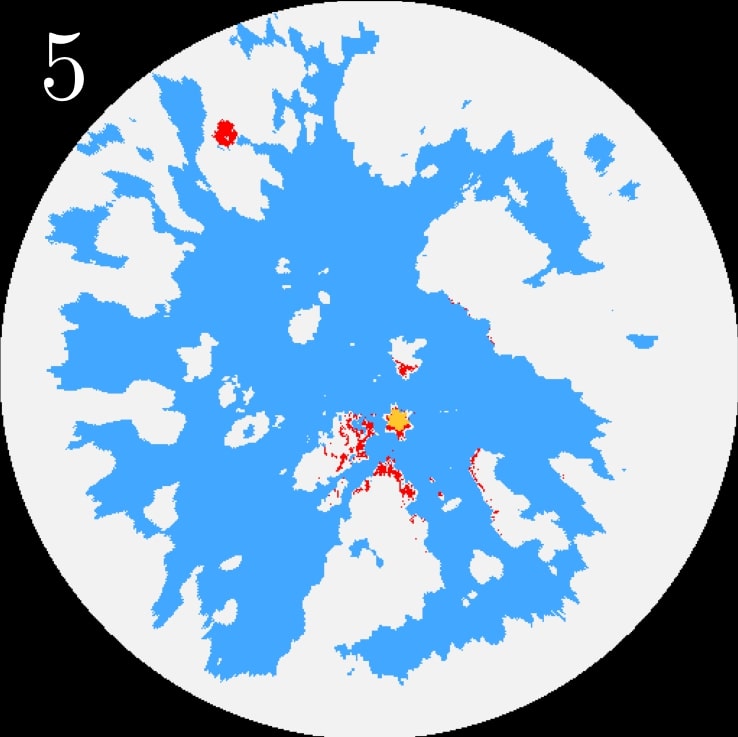}
    \label{fig:sky_image_segmente}
\end{minipage}
\vspace{-1.2\baselineskip}
\caption{From left to right: 1. Raw image of the sky taken with a fish-eye camera, 2. Undistorted image, 3. Localisation of the sun in the image using a sun tracking algorithm~\cite{palettaTemporallyConsistentImagebased2020}, 4. Normalised Blue to Red ratio used in the HYTA algorithm to segment clouds and sky pixels~\cite{liHybridThresholdingAlgorithm2011, hasenbalgBenchmarkingSixCloud2020}, 5. Sky image segmented into 5 classes (sun, clouds, sky, saturation and frame).}
\label{fig:segmentation}
\end{figure*}

{\bf Sky Images} This publicly available dataset was generated at SIRTA's laboratory (France, 48.713° N; 2.208° E, 157 m above average sea level) over 3 years (2017 to 2019)~\cite{sirta}. Images of the sky were taken from the ground with an EKO SRF-02 all-sky camera. Each sample comprises two shots taken with different exposition time: 1/100 sec (long exposure) and 1/2000 sec (short exposure) (Figure~\ref{fig:sky_images}). The short exposure provides more details in the circumsolar area (region close to the sun), whereas the long exposure focuses more on the surrounding sky. From the initial $768 \times 1024$ pixel resolution, sky images are cropped and downscaled to a $128 \times 128$ pixel resolution as suggested in~\cite{sunSolarPVOutput2018}. Images are segmented following the methodology detailed below and illustrated in Figure~\ref{fig:segmentation}.
\vspace{0.5\baselineskip}

The temporal resolution of the sky images is 2-min. Samples with low solar elevation (less than 10° above the horizon) were discarded. The resulting distributions of samples by month, Solar Zenith Angle (SZA) and Global Horizontal Irradiance (GHI) level are given in~\ref{section:dataset_balance} (Figures~\ref{fig:dataset_distribution_month},~\ref{fig:dataset_distribution_sza} and~\ref{fig:dataset_distribution_bins}).

\vspace{0.5\baselineskip}

{\bf Undistortion} Images taken by hemispherical cameras are affected by a strong distortion (Figure~\ref{fig:sky_images}). Consequently, objects located at different angular distances from the main axis of the camera appear differently in the image. In particular, the shape of a cloud grows as it comes closer to the centre of the image. For similar reasons, the trajectory of clouds moving straight appears bent if it does not pass by the axis of the lens. Furthermore, translating the cloud shadow into an irradiance map on the ground for industrial applications requires the distortion of the lens to be taken into account in the modelling.

\vspace{0.5\baselineskip}

To address these difficulties, images are unwrapped assuming an azimuthal equidistant projection. This transformation maintains angular distances relative to the zenith angle. They are projected back onto a plane parallel to the ground (2nd panel in Figure~\ref{fig:segmentation}). For more details see~\cite{palettaTemporallyConsistentImagebased2020}. We performed a comparative study to highlight and discuss the advantages and drawbacks of undistorting images as a preprocessing step for irradiance and cloud map forecasting (Section~\ref{section:undistortion_main}).

\vspace{0.5\baselineskip}

{\bf Sun Position} The angular position of the sun in the sky used to set the condition on the SZA, is included in SIRTA's dataset. However, translating the angular position of the sun into pixel coordinates is difficult because of the fish-eye lens distortion. Therefore, we use here an image-based sun tracking algorithm to estimate the position of the sun in the image~\cite{palettaTemporallyConsistentImagebased2020} (3rd panel in Figure~\ref{fig:segmentation} and Figure~\ref{fig:sun_position_by_day_cleaned_15days_2017}). Implementation details are presented in~\ref{section:sun_tracking}.

\vspace{0.5\baselineskip}

{\bf Image Segmentation} Images are segmented into 5 classes: sky, cloud, sun, saturation and frame. The backbone of the segmentation is the Hybrid Thresholding Algorithm (HYTA)~\cite{liHybridThresholdingAlgorithm2011}. However, its resulting binary classification (sky/clouds) based on a normalised blue to red ratio of the initial image (4th panel in Figure~\ref{fig:segmentation}), lacks reliability in the circumsolar area in specific weather conditions (clear sky or broken sky). To improve the performance of the classification in these conditions, \cite{hasenbalgBenchmarkingSixCloud2020} suggests an adjustment of the classification rule for pixels close to the sun (HYTA+). Here, we decreased the threshold by 10\% for pixels in an area defined by a disk with a diameter of 15 pixels (12\% of the image width) centred on the sun.

\vspace{0.5\baselineskip}

To classify pixels corresponding to the sun, the blue channel of the short exposure image is passed through a Gaussian filter with a standard deviation of half the image width centered on the sun. The threshold to classify the sun is set to 90\% of the maximum pixel intensity of the resulting filtering. Remaining pixels whose blue channel level exceeds 98\% of the maximum theoretical value are classified as saturated. The most common areas of dense saturation are observed when the atmosphere diffuses the light originating from the sun low in the horizon. Finally, black pixels surrounding the image of the sky are classified as frame. The resulting segmentation is depicted in Figure~\ref{fig:segmentation}.

\vspace{0.5\baselineskip}

{\bf Irradiance} The local solar flux predicted by the model is the Global Horizontal Irradiance (GHI). It corresponds to the amount of energy received by a $1\text{m}^2$ surface parallel to the ground over a second (unit: W/$\text{\lowercase{m}}^2$). This scalar quantity used as a supervision signal to train the DL model, is measured on site (Baseline Surface Radiation Network - BSRN Palaiseau) with a ventilated pyranometer (Kipp and Zonen CM22) and reported as its per minute average. Sensors are cleaned from 3 to 5 times a week and are regularly calibrated in Davos by the World Radiation Center.

\subsection{Losses}

The total loss $L_{tot}$ used to trained the network is composed of two components, an irradiance loss ($L_{irra}$) and a segmentation loss ($L_{seg}$) as defined in Equation~\ref{equ:total}. In this study, the weight $\alpha$ is set to 1 or 0 (when $\alpha=0$ the model is not supervised by the prediction of future segmented images).

\vspace{-0.9\baselineskip}

\begin{equation}
  L_{tot} = L_{irra} + \alpha L_{seg}
   \label{equ:total}
\end{equation}

{\bf Future Irradiance Predictions} The loss function for a series of future irradiance predictions $\{\hat{y}_{t+i}\}_{i=1,..,H}$ given corresponding targets $\{y_{t+i}\}_{i=1,..,H}$ is the MSE (Equation~\ref{equ:MSE}).

\vspace{-0.9\baselineskip}

\begin{equation}
  L_{irra} = \frac{1}{H} \sum_{i=1}^{H} (\hat{y}_{t+i} - y_{t+i})^2
   \label{equ:MSE}
\end{equation}

{\bf Future Segmented Images} The segmentation loss $L_{seg}$ is the Cross Entropy. A discount factor $\gamma \in[0,1]$ is used to increase the weight on the first frames of the predicted sequence in the loss function ($\gamma$ was arbitrarily set to 0.9 in this study). Let the segmentation loss at the future timestep $t+i$ be $L_s^{t+i}$ (with $i \in\{1,..,H\}$). The corresponding weighted loss associated with the prediction made at timestep $t$ is:

\begin{equation}
  L_{s} =  \sum_{i=1}^{H} \gamma^i L_s^{t+i}
  \label{equ:loss_discount_factor}
\end{equation}

\subsection{Metrics}

{\bf Forecast Skill} In solar energy, the traditional approach to assess a forecasting model is to evaluate its performance relative to a reference model. The most common baseline model is the smart persistence model (SPM). The persistence model, which predicts unchanged future irradiance levels over a forecast window $\Delta T$, $\hat{y}(t+\Delta T) = y(t)$, is adjusted to take into account diurnal changes of the extra-terrestrial irradiance (Equations~\ref{equ:smart_persistence} and~\ref{equ:smart_persistence_2}). The clear sky index $k_c$ used to model this variability is based on the clear sky irradiance $y_{clr}$ provided by HelioClim~\cite{blancHelioClimProjectSurface2011}. This is a modelled irradiance that is totally exogenous from this study.

\vspace{0.5\baselineskip}

\begin{equation}
   \hat{y}(t+\Delta T) = k_c(t) \, y_{clr}(t+\Delta T)
   \label{equ:smart_persistence}
\end{equation}

\vspace{-0.5\baselineskip}

\begin{equation}
   k_c(t) =  \frac{y(t)}{y_{clr}(t)}
   \label{equ:smart_persistence_2}
\end{equation}

For a given metric error, the Forecast Skill (FS) is computed as presented in Equation~\ref{equ:FS}. The generalisation properties of this metric are used to compare model performances between different datasets and different site locations~\cite{yangVerificationDeterministicSolar2020a}. The score ranges from $-\infty$ to $100\%$. A positive FS indicates an average error smaller than that of the SPM, the closer to $100\%$ the better.

\vspace{-0.5\baselineskip}

\begin{equation}
  \text{FS} = \left(1-\frac{\text{Error}_{forecast}}{\text{Error}_{SPM}}\right) \times 100
   \label{equ:FS}
\end{equation}

As shown by~\cite{vallanceStandardizedProcedureAssess2017}, a major limitation of the FS is its lack of qualitative descriptive ability. For instance, statistical approaches reach high RMSE FS by avoiding the largest errors despite no anticipation skill. This behaviour is notably common with current DL approaches as observed by~\cite{palettaBenchmarkingDeepLearning2021c}. Behaving similarly to a `\textit{very smart persistence}' model, DL models tend to suffer from a recurrent time lag, thus frequently missing critical events such as a cloud hiding the sun.
\vspace{0.5\baselineskip}

{\bf Temporal Distortion Index} To quantify this latency, \cite{frias-paredesIntroducingTemporalDistortion2016} introduced the Temporal Distortion Index (TDI) based on Dynamic Time Warping. The objective of this method is to find the optimal distortion to temporally align a time series with a reference series given a cost function. For a given alignment, the TDI is defined by the fraction of the corresponding distortion relative to the maximum theoretical distortion. Given the type of temporal distortion observed (late or ahead), this indicator can be further decomposed into two components,  in advance ($\text{TDI}_{adv}$) and late ($\text{TDI}_{late}$)~\cite{vallanceStandardizedProcedureAssess2017}.

\vspace{-0.3\baselineskip}

\begin{equation}
  \text{TDI} = \text{TDI}_{adv} + \text{TDI}_{late}
  \label{equ:TDI}
\end{equation}

{\bf 95\% Quantile}. The short horizon of the forecast leads to an unbalance of the dataset towards low changes. Thus, indicators such as the FS based on the averaged errors (RMSE, Mean Absolute Error, etc.) provide little information on the model's ability to predict rare but critical large errors. To evaluate this aspect of the forecasts, we report the 95\% quantile error based on the sorted list of mean absolute errors.

\subsection{Model Training}

The dataset used in the study comprises three years. The first two years (2017 and 2018) compose the training set (180,000 samples). The validation set is composed of 30,000 samples from even day numbers of each month from 2019 (day 2, 4, 6, etc.). The model is tested on 30,000 samples from the remaining odd days from 2019. The TDI metric is evaluated on 200 sequences of 100 consecutive samples (3h20 each) randomly sampled from these test days. The results reported in this paper are averaged over two trainings with different random initialisations.

\section{Results}
\label{section:results}

We compare ECLIPSE with three DL architectures on the 2, 4, 6, 8 and 10-min ahead predictions: ConvLSTM~\cite{palettaBenchmarkingDeepLearning2021c}, TimeSFormer~\cite{bertasiusSpaceTimeAttentionAll2021} and PhyDNet~\cite{leguenDeepPhysicalModel2020a}. TimeSFormer is a convolution-free DL model relying solely on attention mechanisms to extract spatiotemporal features directly from a sequence of frame-level patches. PhyDNet makes predictions constrained by partial differential equations in the latent space. This aims at leveraging physical phenomena such as the cloud cover dynamics, which can be described by advection-diffusion equations. All models take the same input data (RGB sky images) and forecast future irradiance levels. In addition, ECLIPSE and PhyDNet predict the corresponding future segmentations.

\begin{figure}
\centering    
\includegraphics[width=0.48\textwidth]{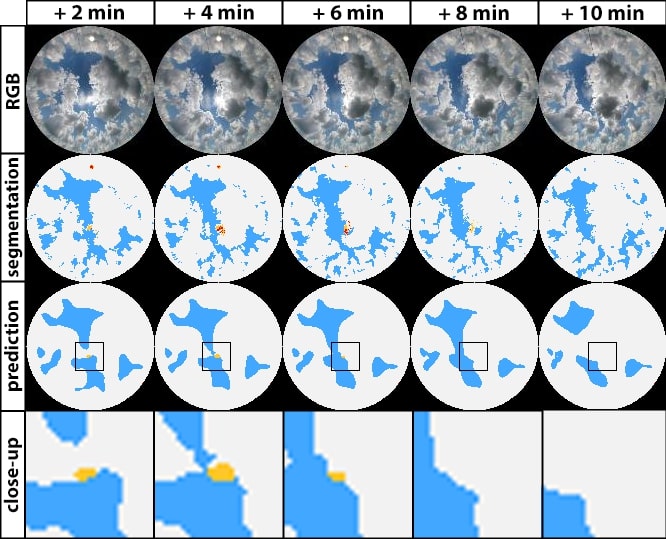}
\caption{From left to right, the 5 future frames (2 to 10-min ahead). From top to bottom: RGB images, target segmentations, predicted segmentations, close-up of the predicted circumsolar area. ECLIPSE predicts the future displacement of the cloud hiding the sun. This sequence corresponds to the sharp irradiance decrease observed around 11:20 in Figure~\ref{fig:series}.}
\label{fig:series4}
\end{figure}

\begin{figure}
\centering
    \includegraphics[width=0.48\textwidth]{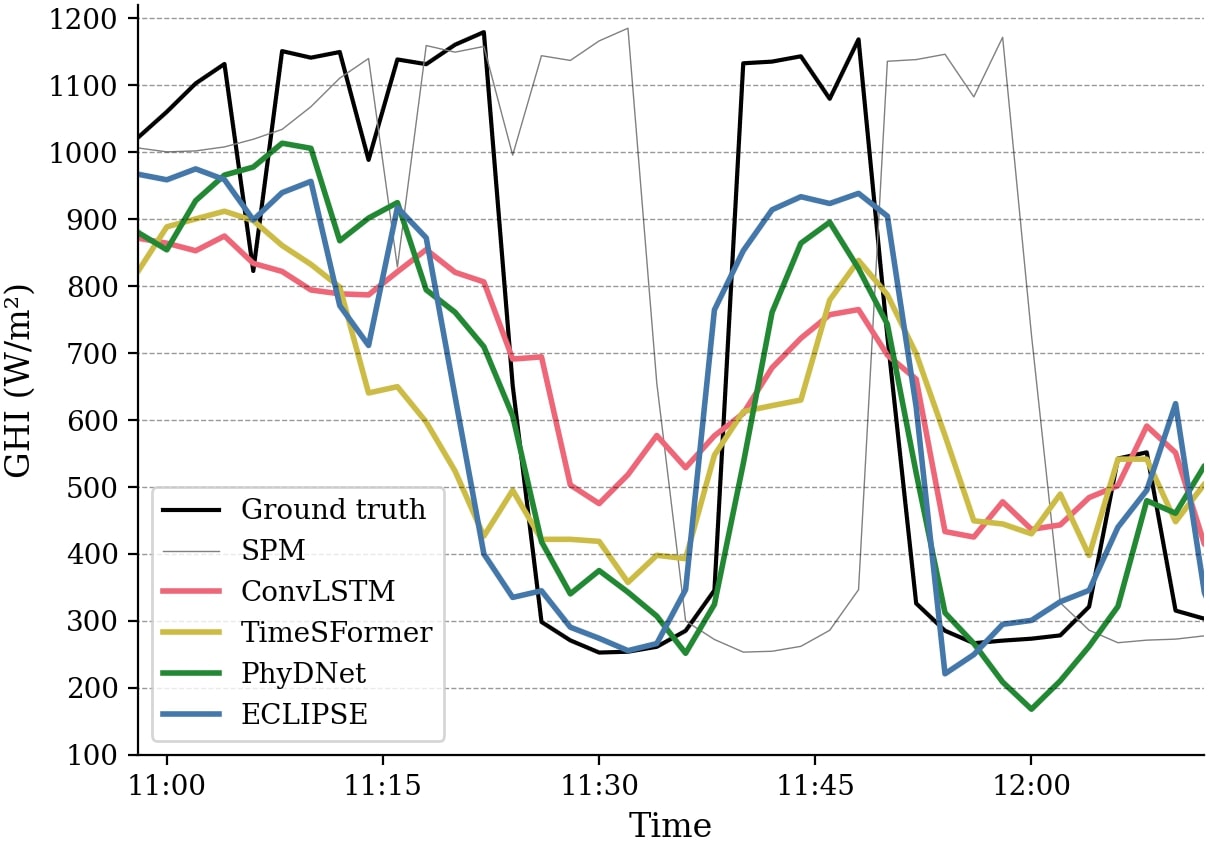}
\vspace{-1.7\baselineskip}  
\caption{10-min ahead irradiance prediction series (test data from 29/05/2019). The sequence of 2 to 10-min ahead predictions in Figure~\ref{fig:series4}, which corresponds to the sharp irradiance fall around 11:20, demonstrates that the model predicts cloud displacement to anticipate the eclipse of the sun by the cloud.}
\label{fig:series}
\end{figure}

\begin{table*}[t]
\begin{center}
\begin{tabular}{lcccccccc}
\hline
 & & \multicolumn{3}{c}{RMSE $\downarrow$ [W/$\text{m}^2$] (Forecast Skill $\uparrow$ [\%])} & & \multicolumn{3}{c}{TDI $\downarrow$ [\%] (Advance / Late)}\\

 Forecast Horizon & $\mid$ & 2-min & 6-min & 10-min & $\mid$ & 2-min & 6-min & 10-min \\
\hline\hline
Smart Pers. && 93.3 (0\%) & 129.0 (0\%) & 143.6 (0\%) && 1.9 (0.0/1.9) & 5.2 (0.0/5.2) & 8.4 (0.0/8.4) \\

ConvLSTM~\cite{palettaBenchmarkingDeepLearning2021c} && 95.6 (-2.5\%) & 107.2 (16.9\%) & 115.7 (19.4\%) && 11.3 (5.7/5.6) & 12.7 (5.7/7.0) & 14.0 (5.9/8.1) \\
TimeSFormer~\cite{bertasiusSpaceTimeAttentionAll2021} && 93.1 (0.2\%) & 105.0 (18.6\%) & 115.2 (19.8\%) && 10.3 (4.8/5.6) & 12.1 (5.1/6.9) & 15.1 (5.9/9.2) \\
PhyDNet~\cite{leguenDeepPhysicalModel2020a} && 87.7 (6.0\%) & 102.0 (20.9\%) & 112.3 (21.8\%) && 10.4 (4.8/5.6) & 11.2 (\textbf{4.7}/6.5) & 13.5 (4.5/9.0) \\
ECLIPSE && \textbf{83.8 (10.2\%)} & \textbf{98.5 (23.6\%)} & \textbf{109.1 (24.0\%)} && \textbf{9.2} (\textbf{4.5/4.7}) & \textbf{11.1} (4.8/\textbf{6.3}) & \textbf{11.9} (\textbf{4.4/7.5}) \\
\hline
\end{tabular}
\end{center}
\vspace{-1.2\baselineskip}
\caption{The models are trained to predict the five next irradiance levels (2, 4, 6, 8 and 10-min ahead). Reported scores corresponding to the 2, 6 and 10-min forecast horizons are averaged over two trainings with different random initialisations. ECLIPSE outperforms other DL models on all horizons, which are increasingly harder to forecast with persistent behaviours.}
\label{tab:results_mean_5horizons}
\end{table*}

\begin{figure*}[ht!]
\centering    
\includegraphics[width=0.92\textwidth]{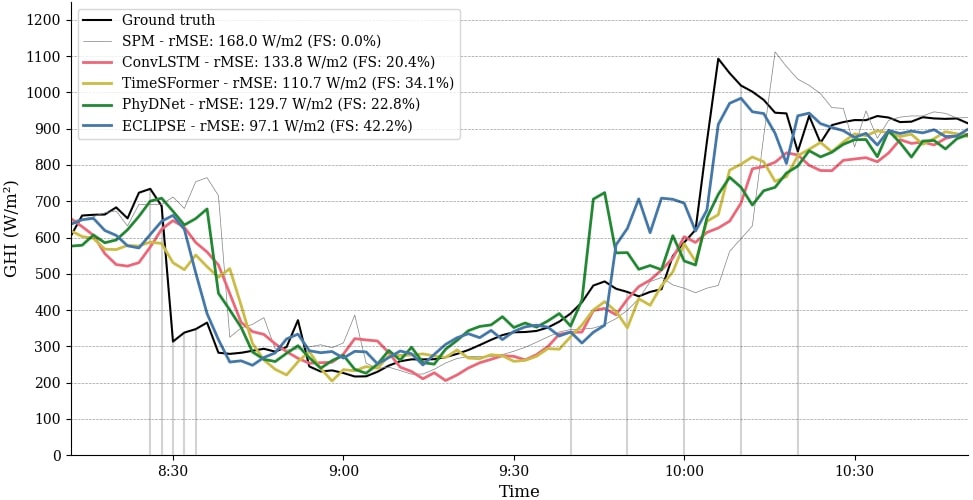}

\vspace{1\baselineskip}

\begin{minipage}[b]{0.19\textwidth}
    \includegraphics[width=1\textwidth]{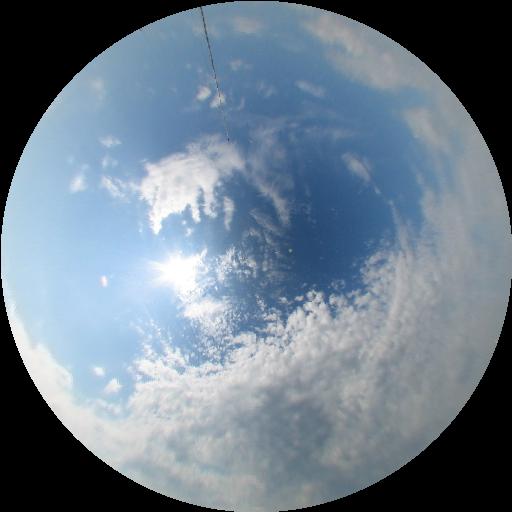} \centering 8:26
  \end{minipage} 
\begin{minipage}[b]{0.19\textwidth}
    \includegraphics[width=1\textwidth]{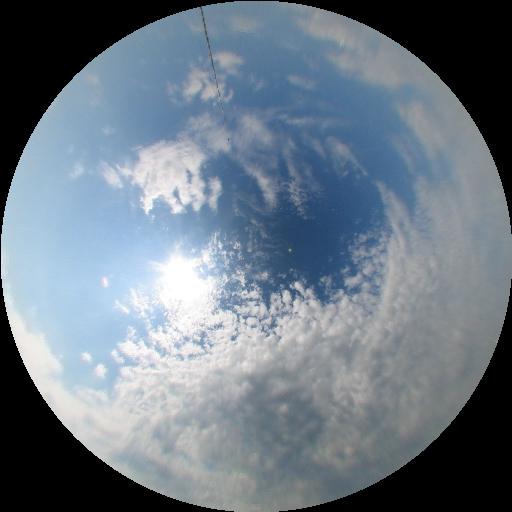} \centering 8:28
  \end{minipage}
  \begin{minipage}[b]{0.19\textwidth}
    \includegraphics[width=1\textwidth]{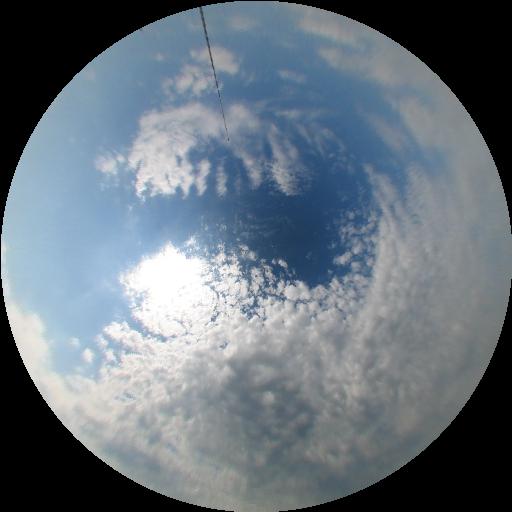} \centering 8:30
  \end{minipage}
   \begin{minipage}[b]{0.19\textwidth}
    \includegraphics[width=1\textwidth]{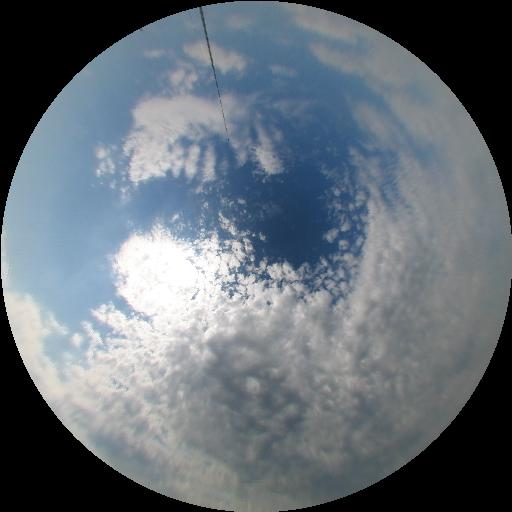} \centering 8:32
  \end{minipage} 
  \begin{minipage}[b]{0.19\textwidth}
    \includegraphics[width=1\textwidth]{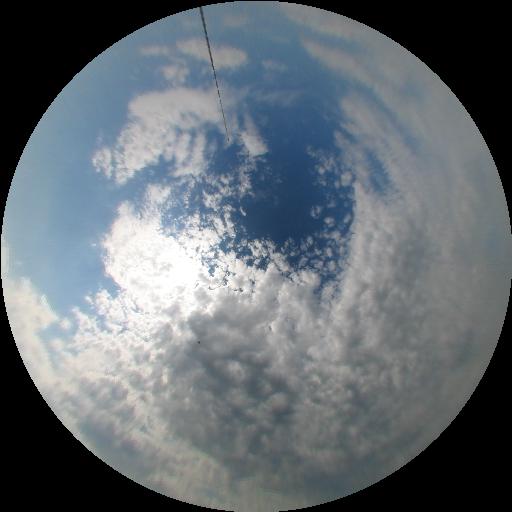} \centering 8:34
\end{minipage}

\vspace{0.5\baselineskip}

\begin{minipage}[b]{0.19\textwidth}
    \includegraphics[width=1\textwidth]{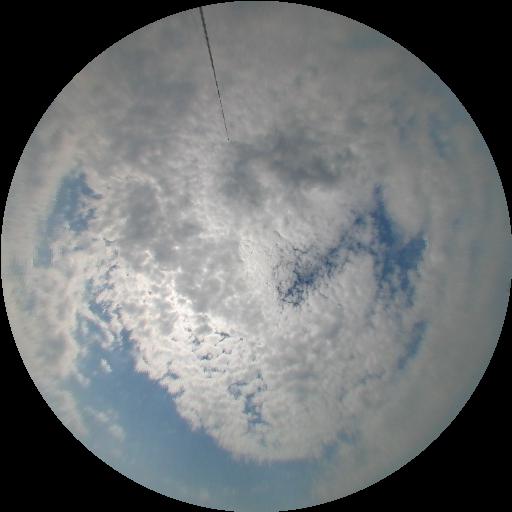} \centering 9:40
  \end{minipage} 
\begin{minipage}[b]{0.19\textwidth}
    \includegraphics[width=1\textwidth]{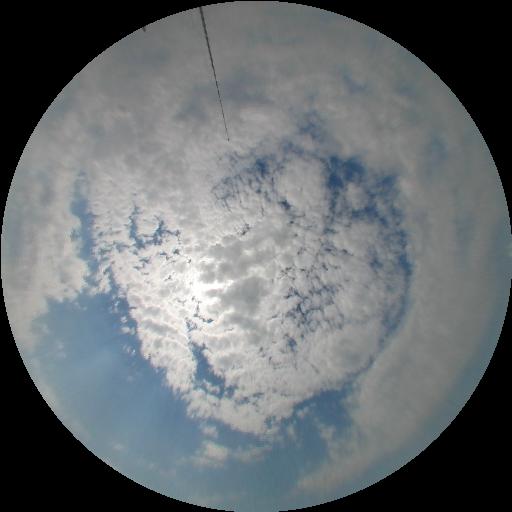} \centering 9:50
  \end{minipage}
  \begin{minipage}[b]{0.19\textwidth}
    \includegraphics[width=1\textwidth]{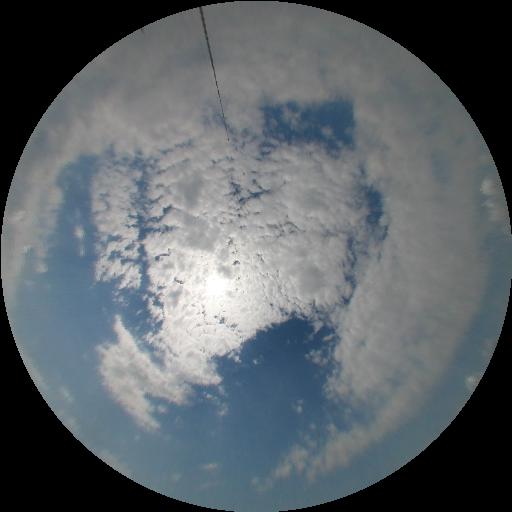} \centering 10:00
  \end{minipage}
   \begin{minipage}[b]{0.19\textwidth}
    \includegraphics[width=1\textwidth]{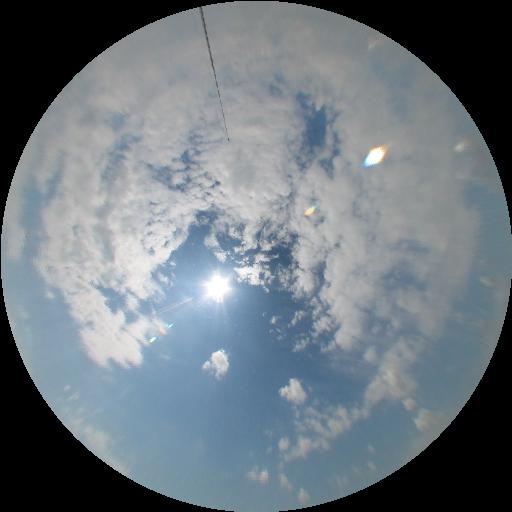} \centering 10:10
  \end{minipage} 
  \begin{minipage}[b]{0.19\textwidth}
    \includegraphics[width=1\textwidth]{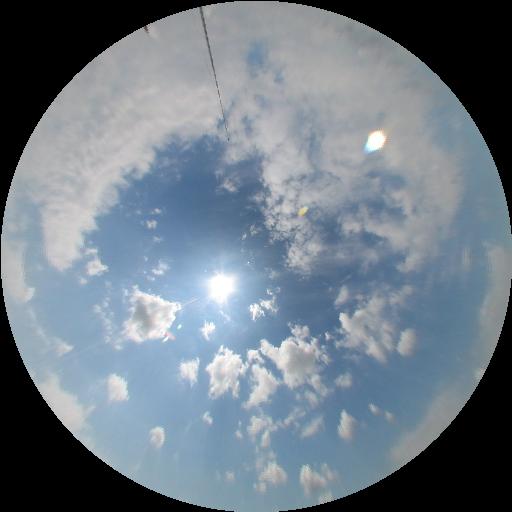} \centering 10:20
\end{minipage}

\caption{10-min ahead prediction curves of all models with corresponding RMSE errors and FS values. ECLIPSE outperforms other models on the FS by a large margin by anticipating both largest ramps around 8:30 and 10:00. Each ramp is illustrated by a sequence of five sky images presented below the graph. Other models suffer from a heavier inertia.}
\label{fig:time_series_benchmark_4_5lf}
\end{figure*}

\subsection{Future State Prediction}
\label{section:future_state_predictions}

By predicting realistic sequences of future segmented images, we show that ECLIPSE can model cloud motion and sun occlusion to forecast critical upcoming events and infer corresponding irradiance changes. As we can see in Figure~\ref{fig:series4}, the model anticipates the eclipse of the sun by the cloud in the middle of the image. This prediction results in an accurate 10-min ahead prediction of the low irradiance value observed in Figure~\ref{fig:series} around 11:30. Figures~\ref{fig:series} and~\ref{fig:time_series_benchmark_4_5lf} show that ECLIPSE better anticipates large irradiance shifts caused by clouds. ECLIPSE and PhyDNet notably display a high variability contrary to the ConvLSTM model, which predicts averaged irradiance levels (Figure~\ref{fig:series}).

\vspace{0.5\baselineskip}

We expect that the longer the forecast window the less reliable the predictions, as the model lacks visual clues to estimate distant cloud shape and displacement. For this reason, the lower the cloud speed the better the ability to anticipate events. In addition, the deterministic short-term nature of cloud dynamics falls along the horizon. This is especially visible in specific meteorological conditions fostering the rapid formation and vanishing of clouds such as altocumulus. In these conditions, probabilistic approaches might provide a more appropriate framework to both explore different futures and support predictions with a confidence level~\cite{sunProbabilisticSolarPower2020} (see Section~\ref{section:model_uncertainty}).

\subsection{Forecasting Performance}

The behaviour difference observed between ECLIPSE and other models regarding temporal distortion, can be further quantified through the TDI. ECLIPSE displays the lowest global temporal distortion on all forecast horizons with improved performances on longer time windows (Table~\ref{tab:results_mean_5horizons}). In particular, distortion linked to late predictions ($\text{TDI}_{late}$) is the lowest for ECLIPSE, which outlines its ability to better anticipate events. Its TDI notably increases with the forecast horizon from 9.2 to 11.9\%. Intuitively, the information contained in past observations might lose predicting value as the forecast horizon grows: clouds which are going to hide the sun in 30-min might not be visible in the present observations. Thus, the longer the prediction window, the more predictions from DL models might resemble those of a statistical model based on trend analysis, i.e. averaged predictions with little anticipation of future events. We therefore expect the TDI scores for all models to converge toward a similar range for longer forecast windows.

\vspace{0.5\baselineskip}

Regarding traditional metrics like the FS, ECLIPSE outperforms the best alternatives on the 2, 6 and 10-min ahead predictions. Whereas some models barely improve over the SPM for the 2-min ahead forecast (ConvLSTM and TimeSFormer), ECLIPSE notably decreases the RMSE from 93.3 to 83.8 W/$\text{m}^2$ (+10.2\% FS). Given that no model has access to past irradiance measurements, this is likely due to ECLIPSE's ability to translate an image of the sky into its corresponding irradiance level.

\begin{table}[t]
\begin{center}
\begin{tabular}{lcccc}
\hline
\noalign{\vskip 0.3mm}
 & & \multicolumn{3}{c}{95\% Quantile [W/$\text{m}^2$]}\\
 Forecast Horizon & $\mid$ & 2-min & 6-min & 10-min\\
\hline\hline
\noalign{\vskip 1mm}
Smart Pers. & & 202.8 & 304.6 & 345.5 \\
\noalign{\vskip 1mm}
ConvLSTM~\cite{palettaBenchmarkingDeepLearning2021c} && 207.9 & 240.1 & 260.0 \\
TimeSFormer~\cite{bertasiusSpaceTimeAttentionAll2021} && 205.4 & 234.0 & 258.7 \\
PhyDNet~\cite{leguenDeepPhysicalModel2020a} && 188.9 & 225.2 & 253.7 \\
ECLIPSE && \textbf{178.1} & \textbf{216.8} & \textbf{245.6} \\
\noalign{\vskip 0.5mm}
\hline
\end{tabular}
\end{center}
\vspace{-1.1\baselineskip}
\caption{Rare event prediction on the 2, 6 and 10-min ahead predictions. Some models perform worse than the SPM on the 2-min ahead forecast.}
\label{tab:rare_events_results_eclipse}
\end{table}

\subsection{Critical Event Prediction}

More than the overall performance, expressed as an RMSE value, anticipating critical events is key in many applications. For instance, hybrid power plants combining solar energy and fossil fuel, rely on an accurate prediction of such events to avoid blackouts caused by the fossil fuel backup not having time to warm-up and compensate a sudden loss of solar power. The 95\% quantile quantifies here the performance of the models on such critical events corresponding to the largest irradiance shifts. ECLIPSE outperforms other models on all horizons, while the SPM notably improves over the ConvLSTM and TimeSFormer models on the 2-min ahead forecasts~\ref{tab:rare_events_results_eclipse}.

\subsection{Ablation Study}

\begin{table*}[ht!]
\begin{center}
\begin{tabular}{lcccccccc}
\hline
\noalign{\vskip 1mm}
 & & \multicolumn{3}{c}{Forecast Skill $\uparrow$ [\%]} & & \multicolumn{3}{c}{TDI $\downarrow$ [\%] (Advance / Late)}\\

Forecast Horizon & $\mid$ & 2-min & 6-min & 10-min & $\mid$ & 2-min & 6-min & 10-min \\
\hline\hline
\noalign{\vskip 1mm}
\noalign{\vskip 1mm}
ECLIPSE && \textbf{10.2\%} & \textbf{23.6\%} & \textbf{24.0\%} && 9.2 (4.5/\textbf{4.7}) & 11.1 (4.8/6.3) & \textbf{11.9} (\textbf{4.4/7.5}) \\
\noalign{\vskip 0.5mm}
- No Temporal Encoder && -2.3\% & 14.2\% & 17.2\% && 9.6 (4.5/5.1) & 14.3 (5.9/8.4) & 16.4 (6.3/10.1) \\
- No Future State Prediction && 6.3\% &  21.1\% & 21.4\% && \textbf{8.9} (\textbf{4.1}/4.8) & \textbf{10.5} (\textbf{4.2}/6.3) & 13.1 (5.1/7.9) \\
- No Segmentation ($\alpha=0$) && 6.4\% & 21.1\% & 21.7\% && 9.4 (4.3/5.1) & 11.0 (5.0/\textbf{6.0}) & 12.7 (4.8/8.0) \\
\noalign{\vskip 1mm}
\hline
\end{tabular}
\end{center}
\vspace{-0.5\baselineskip}
\caption{Ablation study highlighting the relative benefits of ECLIPSE's different modules. The temporal encoder seems to improve the quantitative performances the most while decreasing temporal misalignment. Notably, the video prediction task provides a forecasting gain for all horizons.}
\label{tab:ablation_study}
\end{table*}

We performed an ablation study to quantify the relative benefit of each module of the model (Table~\ref{tab:ablation_study}). We compare ECLIPSE with variants without the temporal encoder; without predicting future states (all future segmentations and irradiance levels are directly decoded from the output of the temporal encoder) and without the segmentation task ($\alpha = 0$). Results show that the temporal module has the largest impact on both temporal misalignment and FS for all forecast horizons. This module performs local (3D and 2D convolutions) and global (3D pooling layers) operations on the 3D tensor representation (times, height, width) generated by the spatial encoder~\cite{huProbabilisticFuturePrediction2020}. Hence, it is better able to capture motion at different scales. This type of temporal architecture has never been applied in the context of solar energy forecasting, and it shows a clear performance improvement over previous studies which use recurrent units such as ConvLSTM.

\vspace{0.5\baselineskip}

Another key finding is that the representation learnt with a cross entropy loss for the segmentation task strongly benefits the regression task. This suggests that the self-supervised signal the model gets from predicting future segmentation greatly benefits cloud motion modelling and acts as a regularizer during training, thereby improving irradiance predictions.

\subsection{Sky Image Undistortion}
\label{section:undistortion_main}

A key preprocessing step in traditional computer vision methods for irradiance forecasting from all-sky cameras involves unwrapping the image taken by the fish-eye lens~\cite{huangCloudMotionEstimation2013, marquezIntrahourDNIForecasting2013a}. To the best of our knowledge, this has never been benchmarked as a preprocessing step in deep learning approaches for irradiance forecasting.

\vspace{0.5\baselineskip}

\begin{figure}
  \centering
  \begin{minipage}[b]{0.23\textwidth}
    \includegraphics[width=\textwidth]{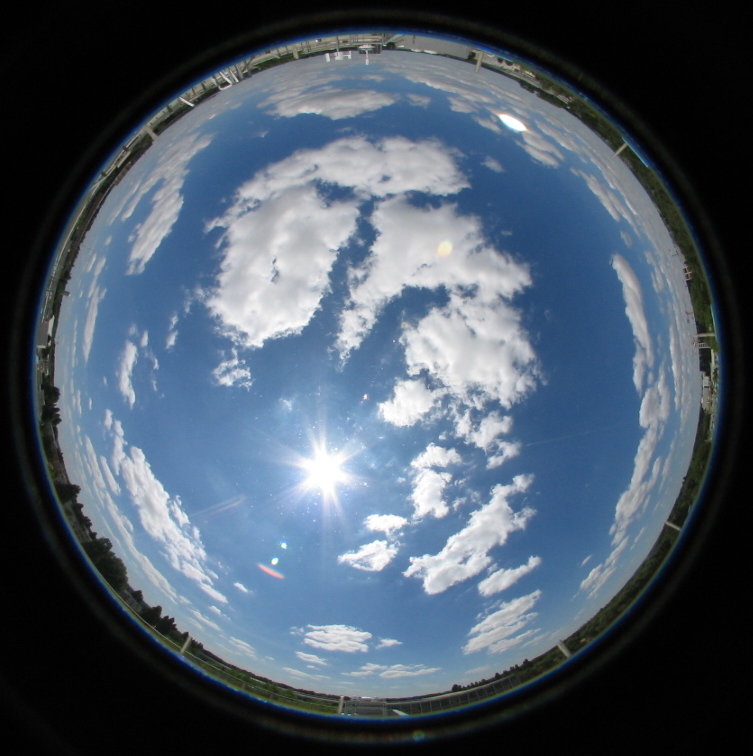}
    \label{fig:20180705132800_01_raw}
  \end{minipage} 
  \begin{minipage}[b]{0.23\textwidth}
    \includegraphics[width=\textwidth]{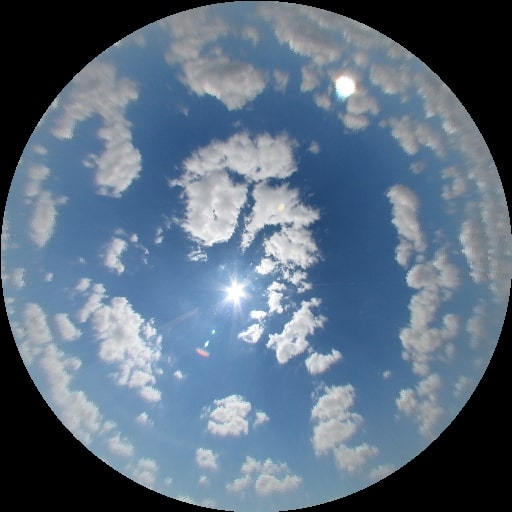}
    \label{fig:20180705132800_01_stereo}
  \end{minipage}
 \caption{Distorted image (left) and the corresponding undistorted image (right) following an equidistant projection. The cloud cover is projected onto a horizontal plane to retrieve cloud shape and trajectory consistency between frames (09/07/2018, 11:04)}
 \label{fig:undistortion_main}
\end{figure}

\begin{table}
\begin{center}
\begin{tabular}{lcccc}
\hline
\noalign{\vskip 1mm}
 & & \multicolumn{3}{c}{Forecast Skill $\uparrow$ [\%]} \\
Forecast Horizon & $\mid$ & 2-min & 6-min & 10-min \\
\hline\hline
\noalign{\vskip 1mm}
\noalign{\vskip 1mm}
ECLIPSE ($\alpha=0$) &&  &  &  \\
\noalign{\vskip 0.5mm}
- Distorted images && \textbf{12.6\%} & \textbf{23.1\%} & \textbf{23.1\%} \\
\noalign{\vskip 0.2mm}
- Undistorted images && 6.4\% & 21.1\% & 21.7\% \\
\noalign{\vskip 1mm}
\hline
\end{tabular}
\end{center}
\vspace{-0.5\baselineskip}
\caption{Comparative experiment to highlight the impact of undistorting sky images prior to feeding them to a deep learning model. The model ECLIPSE is trained to forecast solar irradiance from 2 to 10-min ahead ($\alpha=0$ : no video prediction in this experiment).}
\label{tab:distorted_images_main}
\end{table}

For the sole task of irradiance forecasting, results show that distorted images can be better leveraged by convolutional neural networks than undistorted images (Table~\ref{tab:distorted_images_main}). As can be seen in Figure~\ref{fig:undistortion_main}, the distortion induces the magnifying effect on the central area in the images. Given that the sun remains in that area for a long period of high irradiance (middle of the day), the distortion corresponds to a close-up on the sun. This has been shown to benefit predictions, and in particular very short-term forecasting. \cite{palettaSPIN2021} observed a similar gain from 8 to 14\% (6.4 to 12.6\% here) when training the model on a close-up of the circumsolar area instead of the undistorted image.

\vspace{0.5\baselineskip}

The advantage of undistorted images for industrial applications is a more direct translation, through a projection of the cloud shadow on the ground, of the cloud map into an irradiance map~\cite{nouriEvaluationAllSky2020}. Despite providing less accurate very local forecasts (at the exact location of the camera), the prediction from undistorted images can be better extrapolated to a wider area of several kilometres squared (e.g. a solar farm). In addition, the shape and trajectory consistency as well as the increased area dedicated to distant clouds in undistorted images was shown to benefit longer-term video prediction~\cite{julianPreciseForecastingSky2021}.

\subsection{Forecasts Based on Past Auxiliary data}

In many applications, auxiliary data (e.g. power output, temperature or aerosol concentration) convey valuable information on the operational conditions at the solar site. Better embedding these variables into a computer vision modelling framework would benefit the application of data-driven solar forecasting to industrial sites. For instance, many irradiance forecasting methods rely on time series analysis with the irradiance level as a key variable. Given the high correlation between subsequent measurements, it is notably hard to improve over the persistence models on a very short-term forecasting without knowing past irradiance values. We conducted a set of experiments to highlight the benefit of integrating past irradiance measurements in the modelling and the best approach to do it.

\vspace{0.5\baselineskip}

\ref{section:temporal_lag} highlights that, more than the sun position, it is the past irradiance measurements that strongly benefit very short-term forecasting (from -3.4 to 12.9\% for the ConvLSTM model on the 2-min ahead predictions) and greatly decreases advance temporal distortion, $\text{TDI}_{adv}$ (from 3.8 to 0.9\%). In practice, models using the current irradiance are less likely to predict a trend too early as it would not match the current observed irradiance level. In addition, Table~\ref{tab:results_input_eclipse_main} compares different strategies to integrate past irradiance values for vision-based models (no auxiliary data encoder): adding a fourth channel (RGBI: RGB + Irradiance) or predicting irradiance changes instead of absolute values. Despite the irradiance change prediction strategy providing some performance gains, it is clear that adding a fourth channel with the corresponding irradiance value benefits forecasts the most. Similarly to results obtained in Table~\ref{tab:results_temporal_lag_fs}, very short-term FS jumps from about 10.2 to 18.5\%, while $\text{TDI}$ is more than halved from 9.2 to 4.3\%. \ref{section:past_irradiance} highlights the performance of the four benchmarked models with this new approach (predicting irradiance change from RGBI inputs).

\begin{table*}
\begin{center}
\begin{tabular}{cccccccccc}
\hline
\noalign{\vskip 1mm}
 & & & \multicolumn{3}{c}{Forecast Skill $\uparrow$ [\%]} & & \multicolumn{3}{c}{TDI $\downarrow$ [\%] (Advance / Late)}\\
 \noalign{\vskip 1mm}
 Target & Input Channels & $\mid$ & 2-min & 6-min & 10-min & $\mid$ & 2-min & 6-min & 10-min \\
\hline\hline
\noalign{\vskip 1mm}
Absolute Irradiance & RGB && 10.2\% & 23.6\% & 24.0\% && 9.2 (4.5/4.7) & 11.1 (4.8/6.3) & 11.9 (4.4/7.5) \\
Irradiance Change & RGB && 9.2\% & 23.5\% & 24.4\% && 8.9 (4.1/4.8) & 11.0 (4.7/6.3) & 12.3 (4.6/7.7) \\
Absolute Irradiance & RGBI && 17.6\% & 25.4\% & 25.2\% && 4.3 (1.8/\textbf{2.5}) & 8.6 (3.7/\textbf{4.9}) & 12.3 (4.8/7.6) \\
Irradiance Change & RGBI && \textbf{18.5\%} & \textbf{26.1\%} & \textbf{26.3\%} && \textbf{4.2} (\textbf{1.6}/2.6) & \textbf{7.9} (\textbf{2.9}/5.0) & \textbf{9.5} (\textbf{3.1/6.3}) \\
\noalign{\vskip 1mm}
\hline
\end{tabular}
\end{center}
\caption{Assessment of various strategies to base future predictions on past irradiance measurements. Combining irradiance change forecasting with the RGBI input format provides the best performance for ECLIPSE on both FS and TDI, with the additional irradiance channel accounting for most of the gains.}
\label{tab:results_input_eclipse_main}
\end{table*}

\begin{figure}
\centering    
\includegraphics[width=0.47\textwidth]{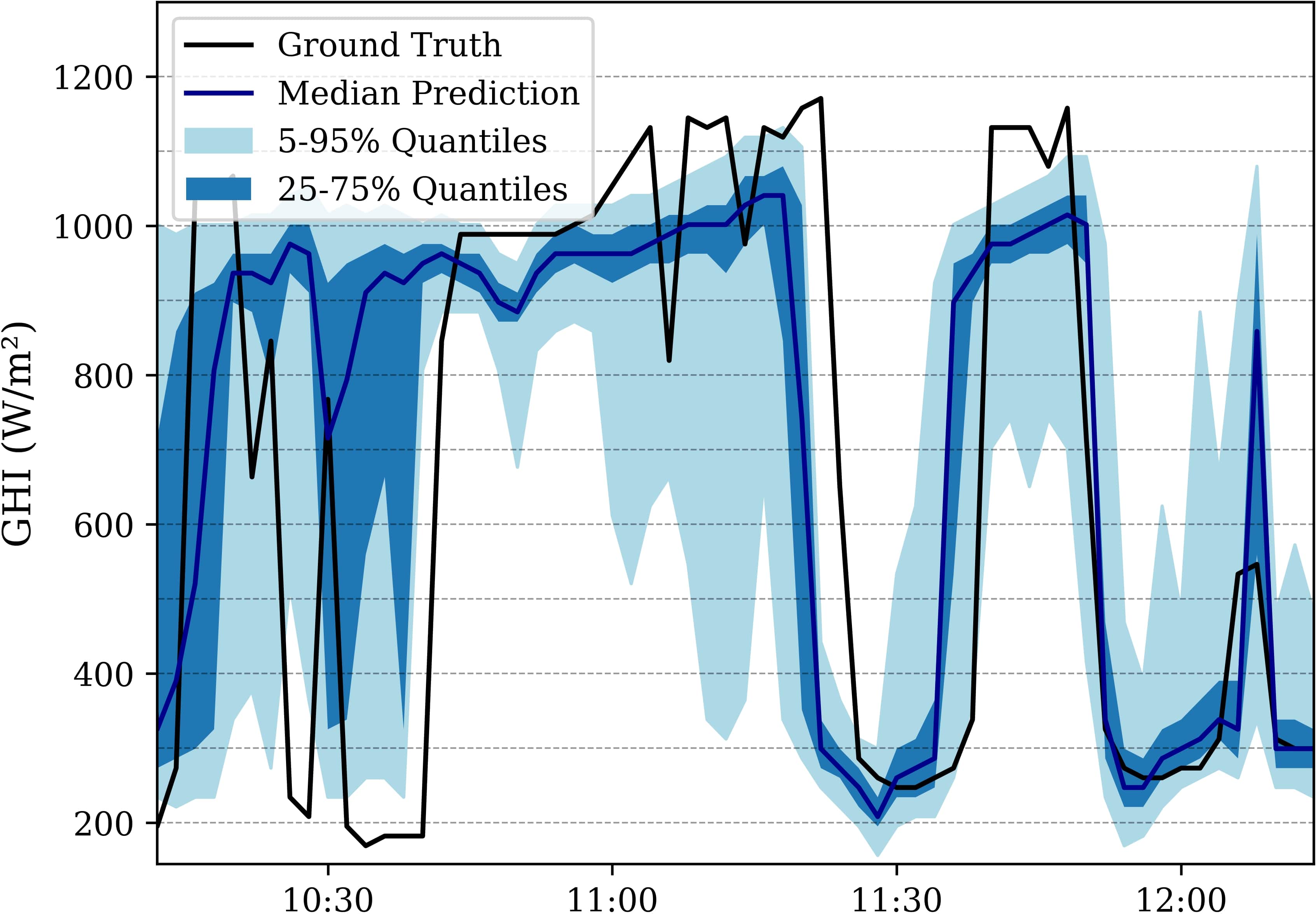}
\vspace{-0.4\baselineskip}
\caption{10-min ahead prediction of the future categorical distribution decoded from future states. The target space is split in 100 equally spaced bins from 0 to 1300 W/$\text{m}^2$. The width of the blue areas illustrates the confidence of the model in its predictions.}
\label{fig:time_series_benchmark_1_5lf_distribution}
\end{figure}

\begin{figure}
     \centering
\begin{subfigure}[b]{0.48\textwidth}
\centering
\begin{minipage}[b]{0.19\textwidth}
    \includegraphics[width=1\textwidth]{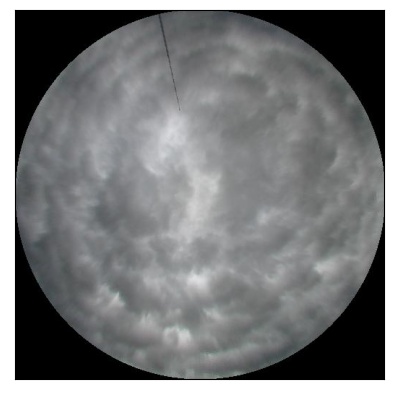}
  \end{minipage} 
\begin{minipage}[b]{0.19\textwidth}
    \includegraphics[width=1\textwidth]{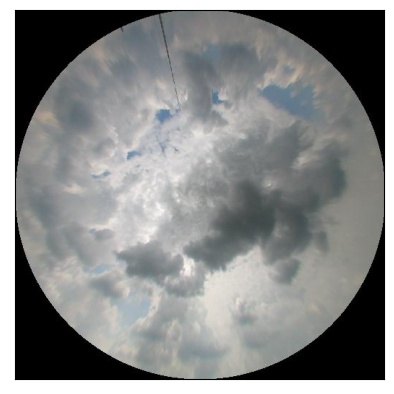}
  \end{minipage}
  \begin{minipage}[b]{0.19\textwidth}
    \includegraphics[width=1\textwidth]{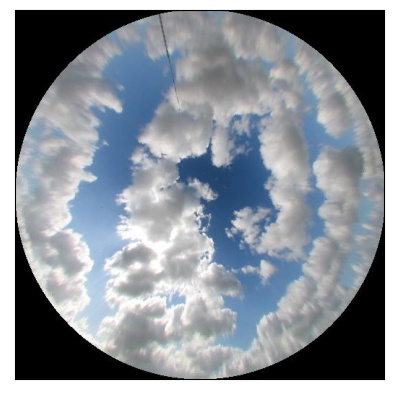}
  \end{minipage}
   \begin{minipage}[b]{0.19\textwidth}
    \includegraphics[width=1\textwidth]{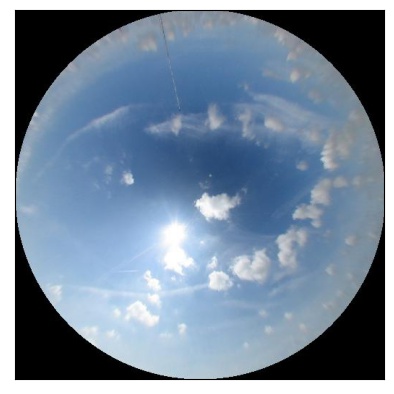}
  \end{minipage} 
  \begin{minipage}[b]{0.19\textwidth}
    \includegraphics[width=1\textwidth]{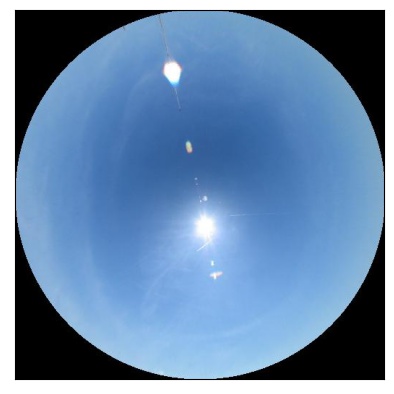}
\end{minipage}
\caption{PC1: extent of the cloud coverage (from left to right: 1C, 2C, 3C, 4C, 5C in the left panel of Figure~\ref{fig:pca_1234_axes_examples}).}
\label{fig:pc1}
\end{subfigure}
\hfill
\vspace{0.05\baselineskip}
\begin{subfigure}[b]{0.48\textwidth}
\centering
\begin{minipage}[b]{0.19\textwidth}
    \includegraphics[width=1\textwidth]{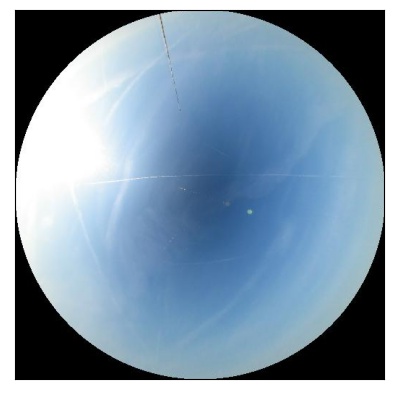}
  \end{minipage} 
\begin{minipage}[b]{0.19\textwidth}
    \includegraphics[width=1\textwidth]{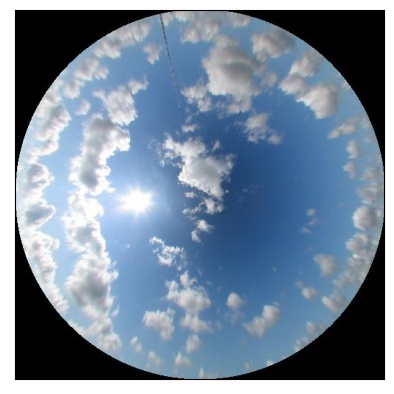}
  \end{minipage}
  \begin{minipage}[b]{0.19\textwidth}
    \includegraphics[width=1\textwidth]{Figs/pca_nb7770_img_201904171036.jpg}
  \end{minipage}
   \begin{minipage}[b]{0.19\textwidth}
    \includegraphics[width=1\textwidth]{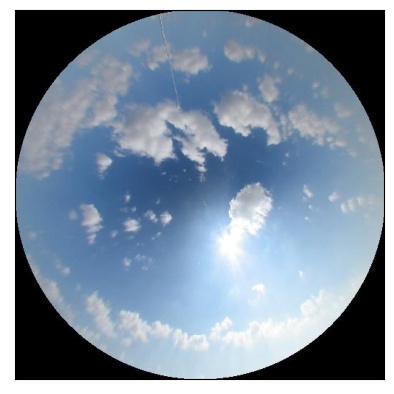}
  \end{minipage} 
  \begin{minipage}[b]{0.19\textwidth}
    \includegraphics[width=1\textwidth]{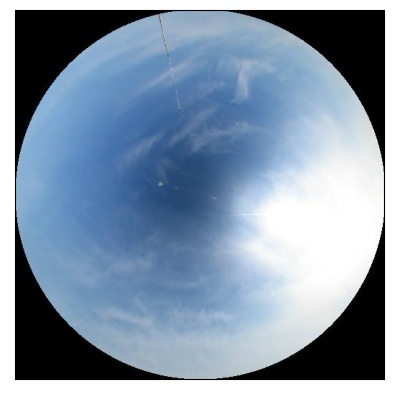}
\end{minipage}
\caption{PC2: horizontal position of the sun in the sky (from left to right: 4A, 4B, 4C, 4D, 4E in the left panel of Figure~\ref{fig:pca_1234_axes_examples}).}
\label{fig:pc2}
\end{subfigure}
\hfill
\vspace{0.05\baselineskip}
\begin{subfigure}[b]{0.48\textwidth}
\centering
\begin{minipage}[b]{0.19\textwidth}
    \includegraphics[width=1\textwidth]{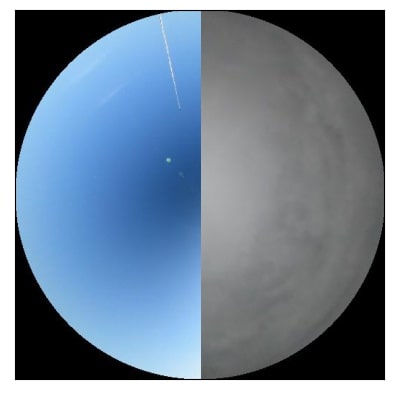}
  \end{minipage} 
\begin{minipage}[b]{0.19\textwidth}
    \includegraphics[width=1\textwidth]{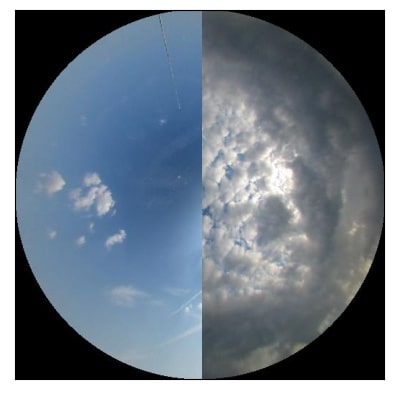}
  \end{minipage}
  \begin{minipage}[b]{0.19\textwidth}
    \includegraphics[width=1\textwidth]{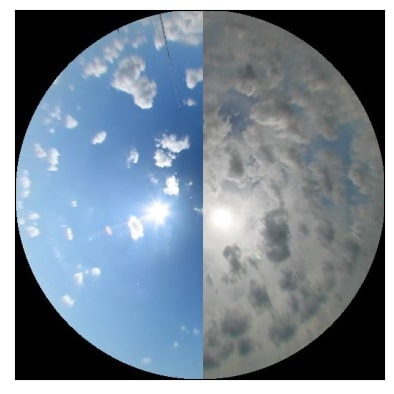}
  \end{minipage}
   \begin{minipage}[b]{0.19\textwidth}
    \includegraphics[width=1\textwidth]{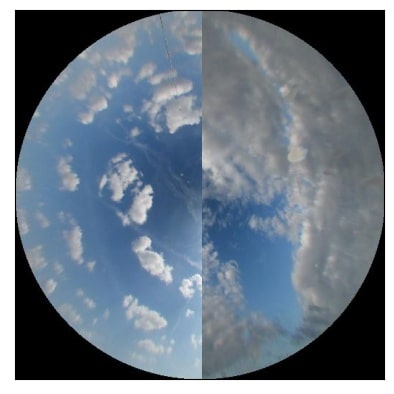}
  \end{minipage} 
  \begin{minipage}[b]{0.19\textwidth}
    \includegraphics[width=1\textwidth]{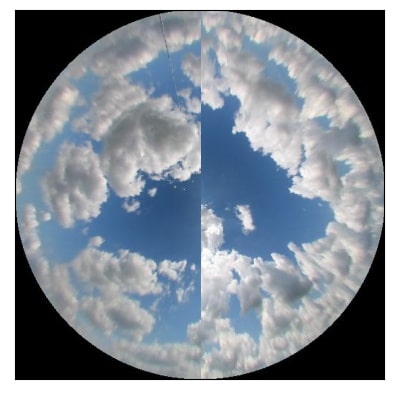}
\end{minipage}
\caption{PC3: spatial variability of the cloud cover: from fully cloudy or fully sunny to partly cloudy. Each image illustrates two neighbouring samples from the distribution (from left to right: 1C, 2C, 3C, 4C, 5C in the right panel of Figure~\ref{fig:pca_1234_axes_examples}).}
\label{fig:pc3}
\end{subfigure}
\hfill
\vspace{0.05\baselineskip}
\begin{subfigure}[b]{0.48\textwidth}
\centering
\begin{minipage}[b]{0.19\textwidth}
    \includegraphics[width=1\textwidth]{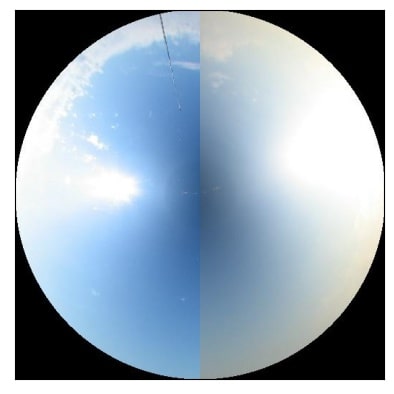}
  \end{minipage} 
\begin{minipage}[b]{0.19\textwidth}
    \includegraphics[width=1\textwidth]{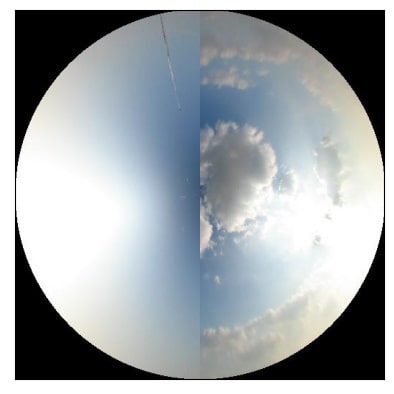}
  \end{minipage}
  \begin{minipage}[b]{0.19\textwidth}
    \includegraphics[width=1\textwidth]{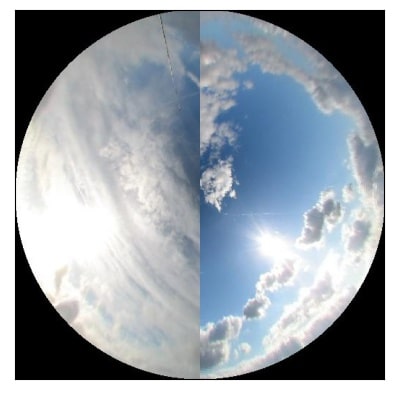}
  \end{minipage}
   \begin{minipage}[b]{0.19\textwidth}
    \includegraphics[width=1\textwidth]{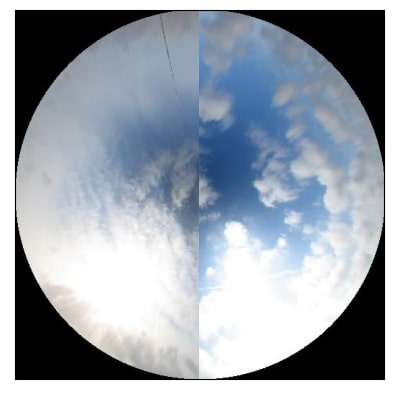}
  \end{minipage} 
  \begin{minipage}[b]{0.19\textwidth}
    \includegraphics[width=1\textwidth]{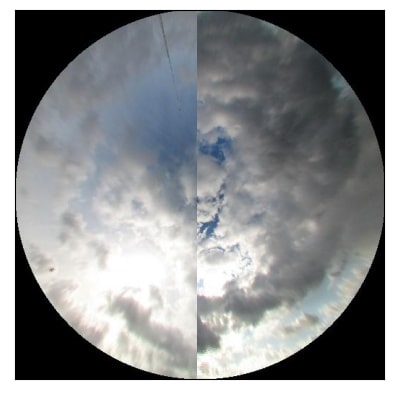}
\end{minipage}
\caption{PC4: vertical position of the sun in the sky. Each image illustrates two neighbouring samples from the distribution (from left to right: 4A, 4B, 4C, 4D, 4E in the right panel of Figure~\ref{fig:pca_1234_axes_examples}).}
\label{fig:pc4}
\end{subfigure}
\vspace{-1\baselineskip}
\caption{Four principal components of the spatio-temporal representation encoded by the model. The variability of each component is illustrated with images draw from the distribution (Figure~\ref{fig:pca_1234_axes_examples}), from low to high values.}
\label{fig:pca_components_examples}
\end{figure}

\begin{figure*}
\centering
\begin{minipage}[b]{0.48\textwidth}
\includegraphics[width=1\textwidth]{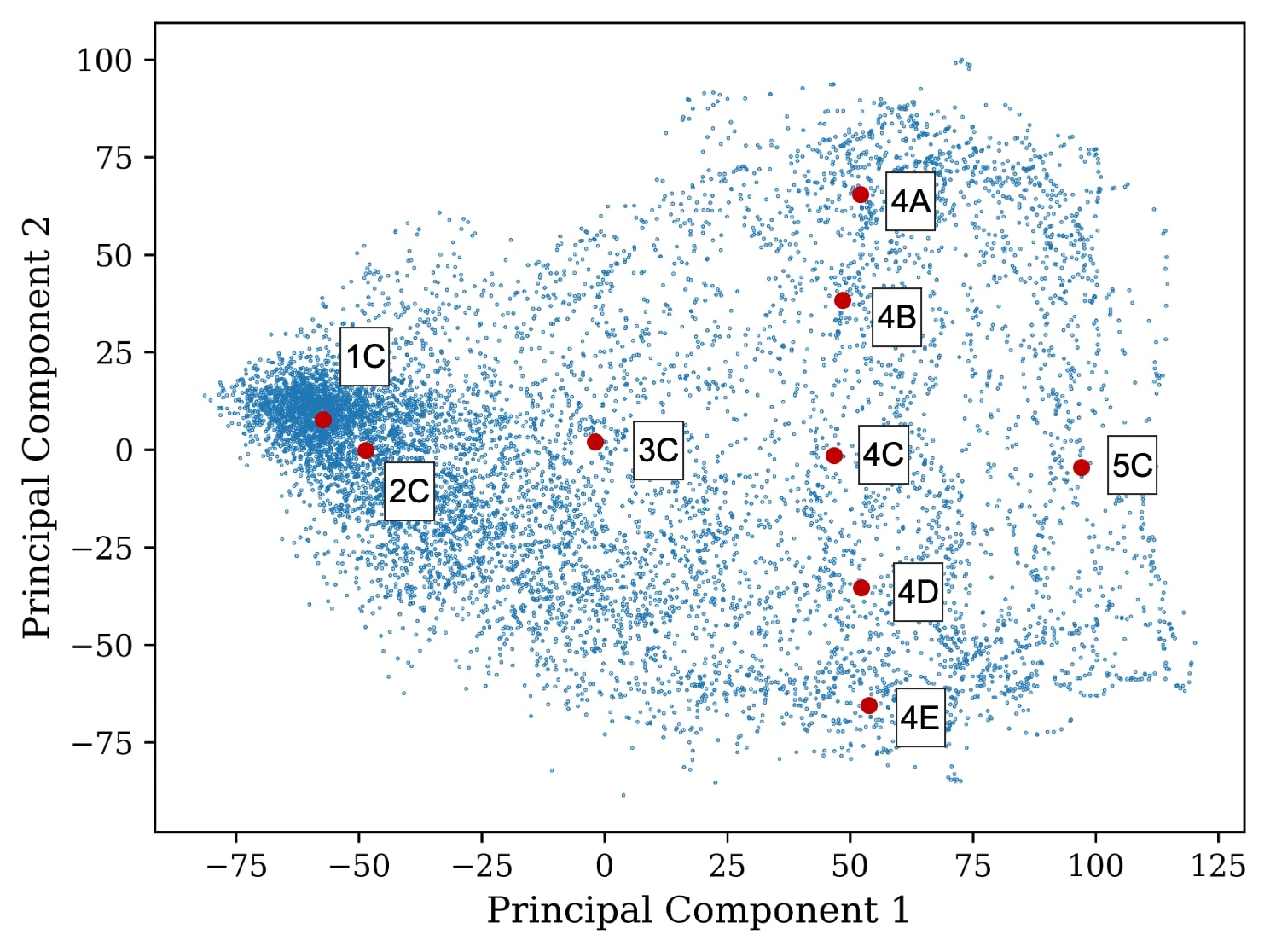}
\end{minipage} 
\begin{minipage}[b]{0.48\textwidth}
\includegraphics[width=1\textwidth]{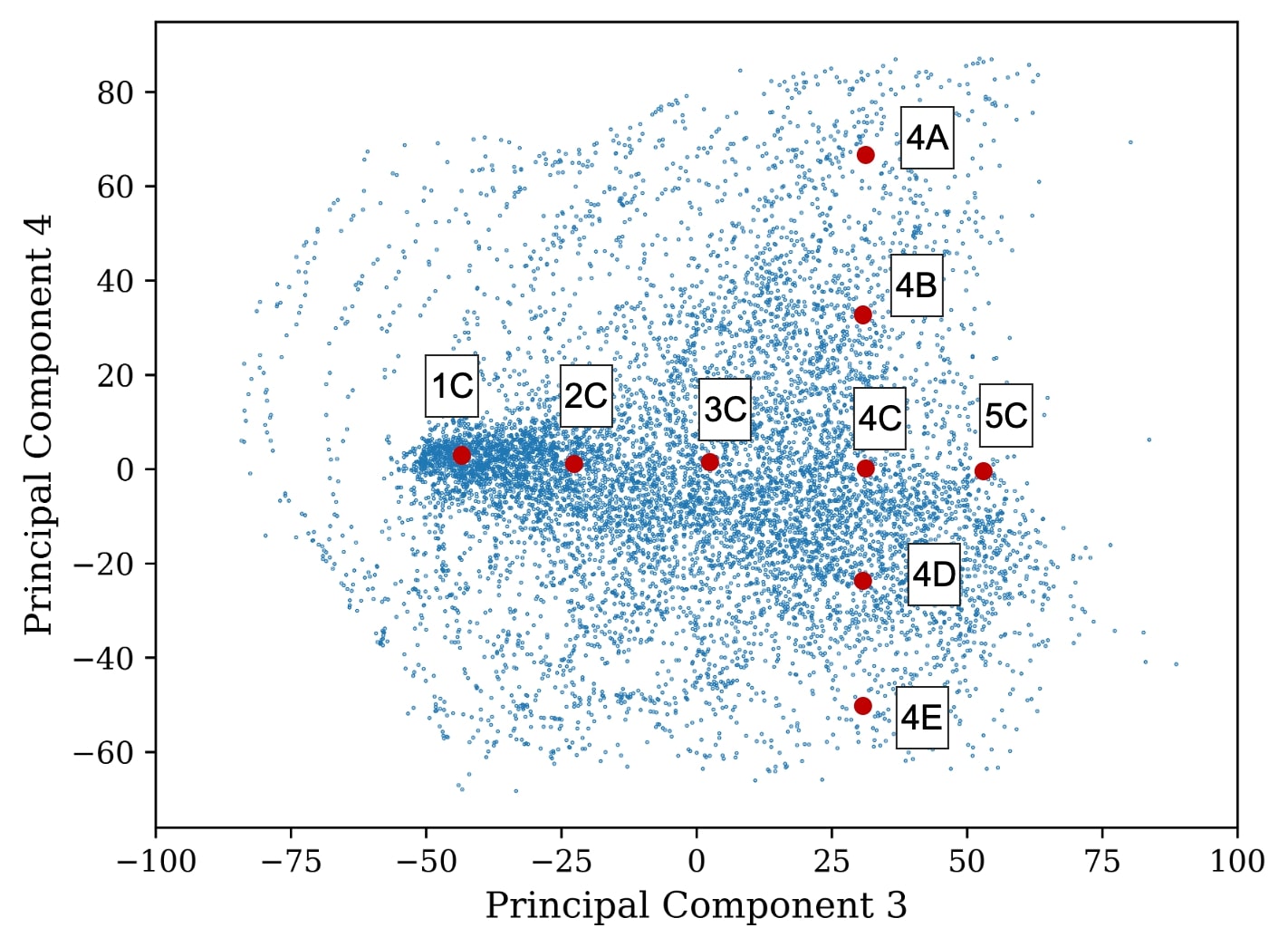}
\end{minipage}
\vspace{-0.5\baselineskip}
\caption{Distribution of the first four principal components of the temporal representation (state $z_t$ in Figure~\ref{fig:model_architecture}) resulting from a principal component analysis on 9000 test samples. The first two components seem to correspond to the extent of cloud coverage, the horizontal position of the sun in the sky, the variability of cloud cover (from a fully cloudy or fully clear sky to partly cloudy sky) and the vertical position of the sun in the sky. To illustrate this point, Figure~\ref{fig:pca_components_examples} shows a few labelled samples drawn from the distribution.}
\label{fig:pca_1234_axes_examples}
\end{figure*}

\subsection{Model Uncertainty}
\label{section:model_uncertainty}

Uncertainty quantification is crucial for many solar applications such as energy trading or hybrid power plant management. Therefore, generating probabilistic forecasts with a neural network~\cite{sonderbyMetNetNeuralWeather2020} facilitates its integration into an operational framework. To illustrate the uncertainty of the model, we add a third decoder similar to the irradiance decoder except for the last layer which outputs a 100 dimensional vector discretizing the output space with 100 equally spaced bins from 0 to 1300 W/$\text{m}^2$. It is trained with the rest of the model by adding a cross entropy loss function to the total loss. The resulting predicted categorical probability distributions are obtained by applying a softmax function to the output of the model. Discrete distributions are more flexible than continuous distributions to approximate complex target distributions (no prior on the shape). In addition, they have been shown to be easier to learn and to produce better performance~\cite{pmlr-v48-oord16}. The width of the dark blue area in Figure~\ref{fig:time_series_benchmark_1_5lf_distribution} shows that the predicted uncertainty changes with the weather variability. At the beginning of the sequence, the rapidly changing cloud cover conditions are associated with a large uncertainty covering the range of irradiance measurements. Notably, the degree of variability observed around 10:30 does not translate into a highly variable median prediction. From 10:40 onward however, the model uncertainty narrows as the predictability in slowly moving conditions increases.

\subsection{Interpretable AI: Visualisation of the Encoded Spatio-temporal Representation}
\label{section:pca_spatiotemporal_features}

To describe the multidimensional representation of past sequences encoded by the model before predicting future states, we perform a principal component analysis (PCA) on the spatio-temporal representation $z_t$ (a $16 \times 16 \times 110$ tensor). The distribution corresponding to the first four principal components (PCs) of the learnt representation is depicted in Figure~\ref{fig:pca_1234_axes_examples}. The respective percentages of the variance explained by these four components are 8.8\%, 4.0\%, 3.1\% and 2.3\%. To further illustrate this PCA, a Gaussian Mixture model was used to cluster the data cloud relative to the two PCs (Figure~\ref{fig:pca_12_clusters_components_examples}).

\vspace{0.5\baselineskip}

The four PCs can be attributed clear visual features (Figure~\ref{fig:pca_components_examples}). PC1 appears to correspond to the extent of cloud coverage, from overcast to clear sky (Figure~\ref{fig:pc1}). PC2 and PC4 account for the horizontal (Figure~\ref{fig:pc2}) and vertical position of the sun in the sky (Figure~\ref{fig:pc4}). PC3 highlights the spatial variability of the cloud coverage on a spectrum from fully cloudy or fully clear (unimodal) of clouds to partly cloudy  (bimodal) (Figure~\ref{fig:pc3}). This spatial variability of the cloud cover correlates with a temporal variability of the irradiance level: the mean standard deviations of the sequence of past irradiance values matching input images in each cluster illustrated in Figure~\ref{fig:pca_components_examples} (c) are 9, 21, 42, 81, 134 W/$\text{m}^2$ respectively. In sky conditions corresponding to a high spatial variability of the cloud cover (see Figure~\ref{fig:series4}), a model with a low anticipation skill might predict values in proximity to the average irradiance level to reduce the risk of a large error due to the effect of a potential sudden event. Conversely, ECLIPSE would try to actively predict the future event on time from other PCs ($\geqslant$ 5).

\vspace{0.5\baselineskip}

Despite ECLIPSE's apparent ability to better extract temporal features from the data, all four principal components of the spatio-temporal representation does not seem to correspond to temporal, but mostly spatial features (position of the sun, extent of the cloud cover). Even the spatial variability of the cloud cover (PC3), which could be linked to the temporal variability of the irradiance level in these sky conditions, can be considered as a spatial feature as it could be extracted from a single image. One interpretation of this is that the standard problem of irradiance forecasting, specifically the minimisation of a mean error (RMSE, MAE, etc.) without considering temporal misalignment, is currently in large part tackled without a reliable modelling of cloud motions (although notably in a smaller proportion with ECLIPSE). This kind of causal confusion might explain why models based on spatial perception with low skills in temporal modelling (a CNN model without recurrent units for instance) perform similarly to ECLIPSE on the RMSE metric. In addition, this could account for a CNN architecture's ability to score a FS as high as 25.47\% on the 10 to 20-min ahead averaged GHI prediction solely from a single image of the sky~\cite{fengSolarNetSkyImagebased2020a} on another dataset~\cite{fengOpenSolarPromotingOpenness2019}. Although temporal features are crucial for the key objective of vision-based solar energy forecasting, current models seem to predominantly rely on spatial features to improve over the SPM. This emphasises the need to improve the techniques with architectures, like ECLIPSE, better able to extract temporal features from the data.

\begin{figure*}[ht!]
\centering
\begin{minipage}[b]{0.3\textwidth}
    \includegraphics[width=1.0\textwidth]{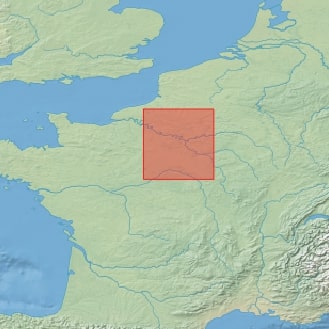}
    \label{fig:satellite_region}
  \end{minipage} 
  \quad
  \begin{minipage}[b]{0.3\textwidth}
    \includegraphics[width=1.0\textwidth]{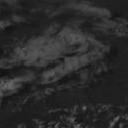}
    \label{fig:satellite_image}
  \end{minipage}
    \quad
  \begin{minipage}[b]{0.3\textwidth}
    \includegraphics[width=1.0\textwidth]{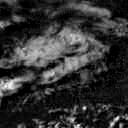}
    \label{fig:cloud_index_map}
  \end{minipage}
\vspace{-0.1\baselineskip}
\caption{From left to right: area covered by the satellite images, greyscale satellite image~\cite{eumetsatorganizationEUMETSATEuropeanOrganisation1991} and its corresponding cloud index map~\cite{palettaSPIN2021} centred on SIRTA's Laboratory (48.713° N, 2.208° E)~\cite{sirta} and acquired the 1 July 2018 at 16:00 UTC. Notably, the right image has a higher contrast compared to the raw observation (middle panel) which was collected in low lighting conditions.}
\label{fig:satellite_images}
\end{figure*}

\begin{table*}[ht!]
\begin{center}
\begin{tabular}{lcccccccc}
\hline
 & & \multicolumn{3}{c}{RMSE $\downarrow$ [W/$\text{m}^2$] (Forecast Skill $\uparrow$ [\%])} & & \multicolumn{3}{c}{95\% Quantile $\downarrow$ [W/$\text{m}^2$]}\\

 Forecast Horizon & $\mid$ & 10-min & 30-min & 50-min & $\mid$ & 10-min & 30-min & 50-min \\
\hline\hline
\noalign{\vskip 1mm}
Smart Pers. && 110.9 (0\%) & 136.0 (0\%) & 145.4 (0\%) && 255.3 & 319.7 & 340.5 \\
ConvLSTM~\cite{palettaBenchmarkingDeepLearning2021c} && 101.1 (8.8\%) & 121.9 (10.4\%) & 132.7 (8.7\%) && \textbf{235.5} & 279.6 & 299.7 \\
TimeSFormer~\cite{bertasiusSpaceTimeAttentionAll2021} && \textbf{99.4 (10.4\%)} &  120.3 (11.5\%) & 130.4 (10.3\%) && 237.4 & 275.3 & 297.0 \\
PhyDNet~\cite{leguenDeepPhysicalModel2020a} ($\alpha=0$)  && 101.8 (8.2\%) & 122.0 (10.3\%) & 132.3 (9.0\%) && 239.6 & 281.1 & 301.6 \\
ECLIPSE ($\alpha=0$) && 101.0 (8.9\%) & \textbf{117.0 (14.0\%)} & \textbf{125.7 (13.5\%)} && 235.7 & \textbf{268.4} & \textbf{290.7} \\
\noalign{\vskip 1mm}
\hline
\end{tabular}
\end{center}
\vspace{-0.5\baselineskip}
\caption{The models are trained to predict the next five irradiance levels (10, 20, 30, 40 and 50-min ahead). Reported scores corresponding to the 10, 30 and 50-min forecast horizons are averaged over two trainings.}
\label{tab:results_eumetsat_data_closeup}
\end{table*}

\section{Irradiance forecasting from satellite images}
\label{section:eumetsat}

For longer range forecasts (30-min to 2h), all-sky cameras lack vision of distant clouds to anticipate future events. This can be done using satellite imagery as an observation tool of the cloud cover dynamics over a large area. For this application, i.e. vision-based irradiance forecasting from satellite images, machine learning models benefit from short processing time compared to more extensive numerical weather models~\cite{perezDeepLearningModel2021b, siPhotovoltaicPowerForecast2021a, nielsenIrradianceNetSpatiotemporalDeep2021}.

\vspace{0.5\baselineskip}

We use in this experiment, a set of greyscale satellite images collected by the Spinning Enhanced Visible and Infrared Imager (SEVIRI) on board the geostationary satellite Meteosat Second Generation (MSG) operated by EUMETSAT~\cite{eumetsatorganizationEUMETSATEuropeanOrganisation1991} (Meteosat SEVIRI Rapid Scan image data~\footnote{https://navigator.eumetsat.int/product/EO:EUM:DAT:MSG:MSG15-RSS}). These $128 \times 128$ pixel images cover a $2.2^{\circ}  \text{(latitude)}\times2.2^{\circ}$ (longitude) area centred on SIRTA's laboratory (48.713° N, 2.208° E). The dataset comprises 46000 training samples from 2017/2018, 9000 validation and 9000 test samples from 2019 with a 5-min temporal resolution. Validation and test samples are taken from distinct days.

\vspace{0.5\baselineskip}

Similarly to Heliosat methods~\cite{SolarEnergyAssessment2003, rigollierMethodHeliosat2Deriving2004}, a cloud index (or “cloud albedo”~\cite{muellerNewAlgorithmSatelliteBased2012}) is derived from the calibrated radiance observed by the satellite sensor. Following the method presented in~\cite{palettaSPIN2021}, the cloud index (CI) is computed from the statistics of the pixel values with $\rho(i, j, t)$ the value of a given pixel ($i,j$) at time $t$; $\rho_{min}(i, j, t-N:t)$ the minimum pixel value of the same pixel ($i,j$) at the same time $t$ of the day over the last $N$ days ($N=10$ here) and $\rho_{max}(t)$ the maximum pixel value in the image at time $t$ (Equation~\ref{equ:cloud_index}). This processing step better reveals the spatial distribution of the cloud cover without being affected by the diurnal lighting condition changes (see Figure~\ref{fig:satellite_images}).

\begin{equation}
   \text{CI} \; (i, j, t) = \frac{\rho(i, j, t)- \rho_{min}(i, j, t-N:t)}{\rho_{max}(t) - \rho_{min}(i, j, t-N:t)}
   \label{equ:cloud_index}
\end{equation}

\vspace{0.5\baselineskip}

In the same way as~\cite{palettaSPIN2021}, we trained deep learning models to forecast 10 to 50-min ahead future irradiance levels at a 10-min resolution from past satellite observations (Figure~\ref{fig:satellite_images}) taken at time $t$, $t - 10\text{-min}$, $t-20\text{-min}$, $t-30\text{-min}$ and $t-40\text{-min}$. Analogously to sky image-based solar forecasting, the input context of models is a set of $128 \times 128 \times 3$ blocks composed of three channels each: the raw satellite image, its corresponding cloud index map and an irradiance channel indicating the level of solar radiation at the time of observation. Irradiance values correspond to the same 1-min average measurements collected at Sirta's lab and presented in Section~\ref{section:dataset}, averaged here over a 5-min time window (e.g. the 30-min ahead prediction corresponds to the 26 to 30-min measurements average). We report the performance of each model based on their 10, 30 and 50-min ahead predictions (see Table~\ref{tab:results_eumetsat_data_closeup}).

\vspace{0.5\baselineskip}

The results show that all models outperform the SPM by around 10-15\% based on the FS metric. In particular, ECLIPSE improves over other models on the 30 and 50-min ahead predictions regarding the FS metrics. Additionally, it is better at anticipating events corresponding to larger irradiance changes as indicated by the 95\% quantiles. On the shortest horizon however, the TimeSFormer model is the best performing. Its finer analysis of spatial features in the last image of the sequence, which best correlate with the current solar resource availability, is likely due to the direct application of attention mechanisms on uncompressed image patches. On the contrary, convolution-based architectures start by compressing images spatially, hence some of the information on the exact spatial localisation of clouds is lost in deeper layers of the network.

\vspace{-0.3\baselineskip}

\section{Conclusion}
\label{section:conclusion}

\vspace{-0.2\baselineskip}

To facilitate operational solar forecasting, we presented a spatio-temporal network architecture able to predict plausible future segmented images and corresponding irradiance levels and associated uncertainties from a sequence of sky images or satellite images. A new segmentation method to determine sun masks was presented to extend the traditional binary (cloud/sky) segmentation approach, hence facilitating the translation of a predicted cloud map to the corresponding local irradiance map or solar panel occlusion map. The proposed architecture breaks the \textit{persistence barrier} by considerably decreasing time lag, while outperforming other models on critical event prediction. Different strategies to integrate auxiliary meteorological or operational variables are exposed though ablation studies. For instance, past irradiance values are shown to strongly benefit short-term forecasting while limiting early temporal misalignment. On the sole objective of irradiance forecasting, the distortion induced by the camera was shown to improve predictions by providing more details on the region of the sky directly above the camera. To interpret the information generated by the DL model, an explainable AI approach was applied. The principal components of the learned representation encoding the past sequence of sky images revealed that the importance of spatial features outweigh that of temporal features in current video-based irradiance forecasting approaches.

\vspace{0.5\baselineskip}

This study indicates several research directions to further improve the techniques of short-term irradiance forecasting based on computer vision: evaluating in detail the impact of segmentation strategies on irradiance predictions (e.g. number of classes, cloud type differentiation), refining the segmentation of sky images with infrared observations~\cite{ajithDeepLearningBased2021} or making the training self-supervised by predicting RGB images, adapting the learning strategy to an imbalanced dataset, or better modelling spatio-temporal feature interactions through attention mechanisms. For sky image-based irradiance forecasting, the appropriate region of the image corresponding to clouds shadowing a nearby solar site could be directly given to the model in a similar approach as~\cite{siPhotovoltaicPowerForecast2021a} with satellite imagery. Finally, the applicability of the trained model to another site is an open key question requiring further inquiry.

\vspace{1\baselineskip}

{\bf Acknowledgements} The authors acknowledge SIRTA for providing the data used in this study. We are grateful to Gisela Lechuga, Aleksandra Marconi, Julian Nappert, Jordi Badosa, Philippe Blanc and Marcos Gomes-Borges for their technical assistance and valuable comments on the manuscript. This research was supported by ENGIE Lab CRIGEN, EPSRC and the University of Cambridge.

{\small
\bibliographystyle{elsarticle-harv}
\bibliography{MyLibrary}
}

\newpage

\appendix

\renewcommand\thefigure{\thesection.\arabic{figure}}
\renewcommand\thetable{\thesection.\arabic{table}}

\onecolumn
\section{Dataset Balance}
\label{section:dataset_balance}
\setcounter{figure}{0}
\setcounter{table}{0}

\begin{figure}[ht!]
\centering
\begin{minipage}[b]{0.31\textwidth}
    \includegraphics[width=1\textwidth]{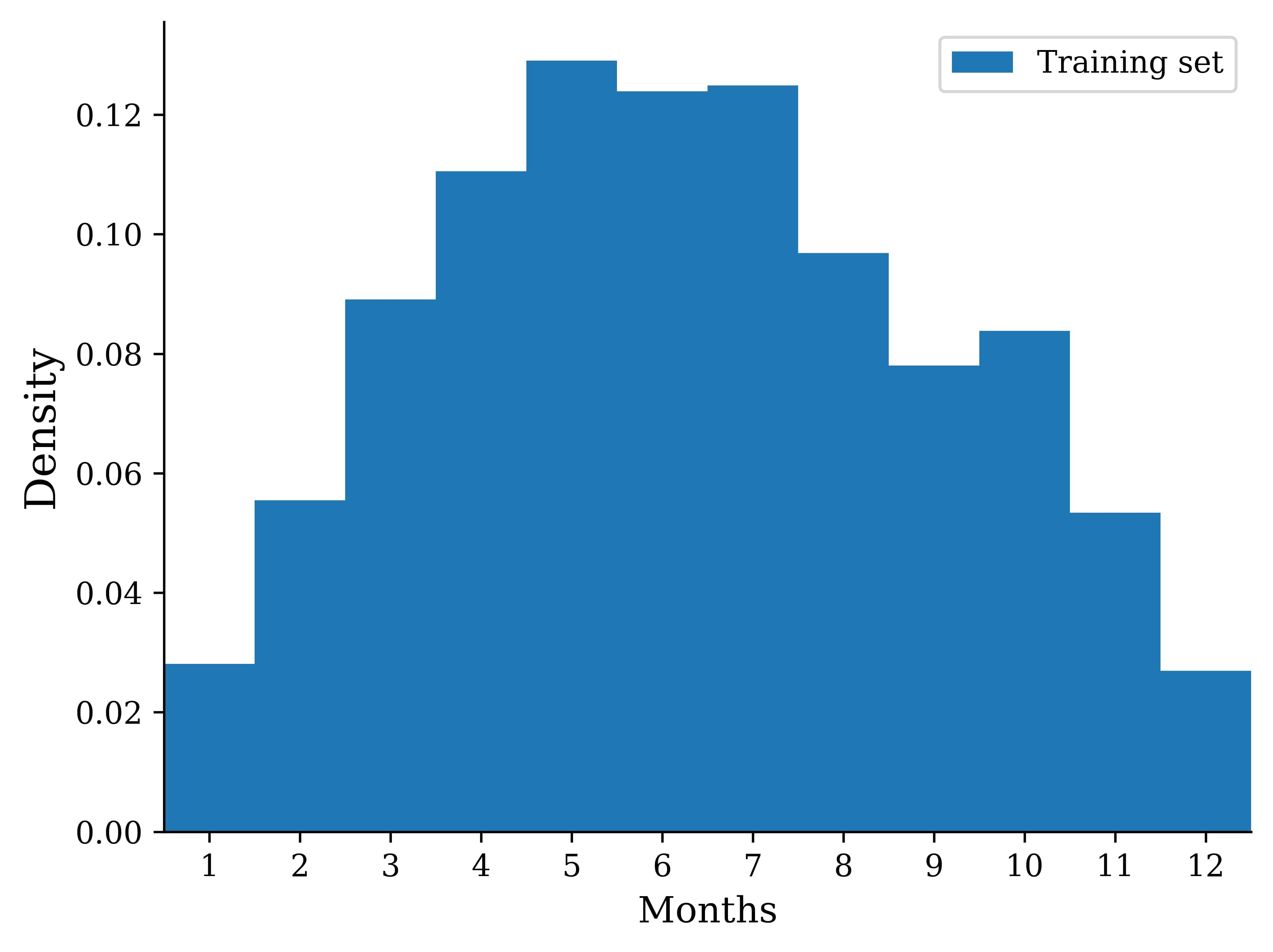}
    \label{fig:hist_months_train}
  \end{minipage} 
  \begin{minipage}[b]{0.31\textwidth}
    \includegraphics[width=1\textwidth]{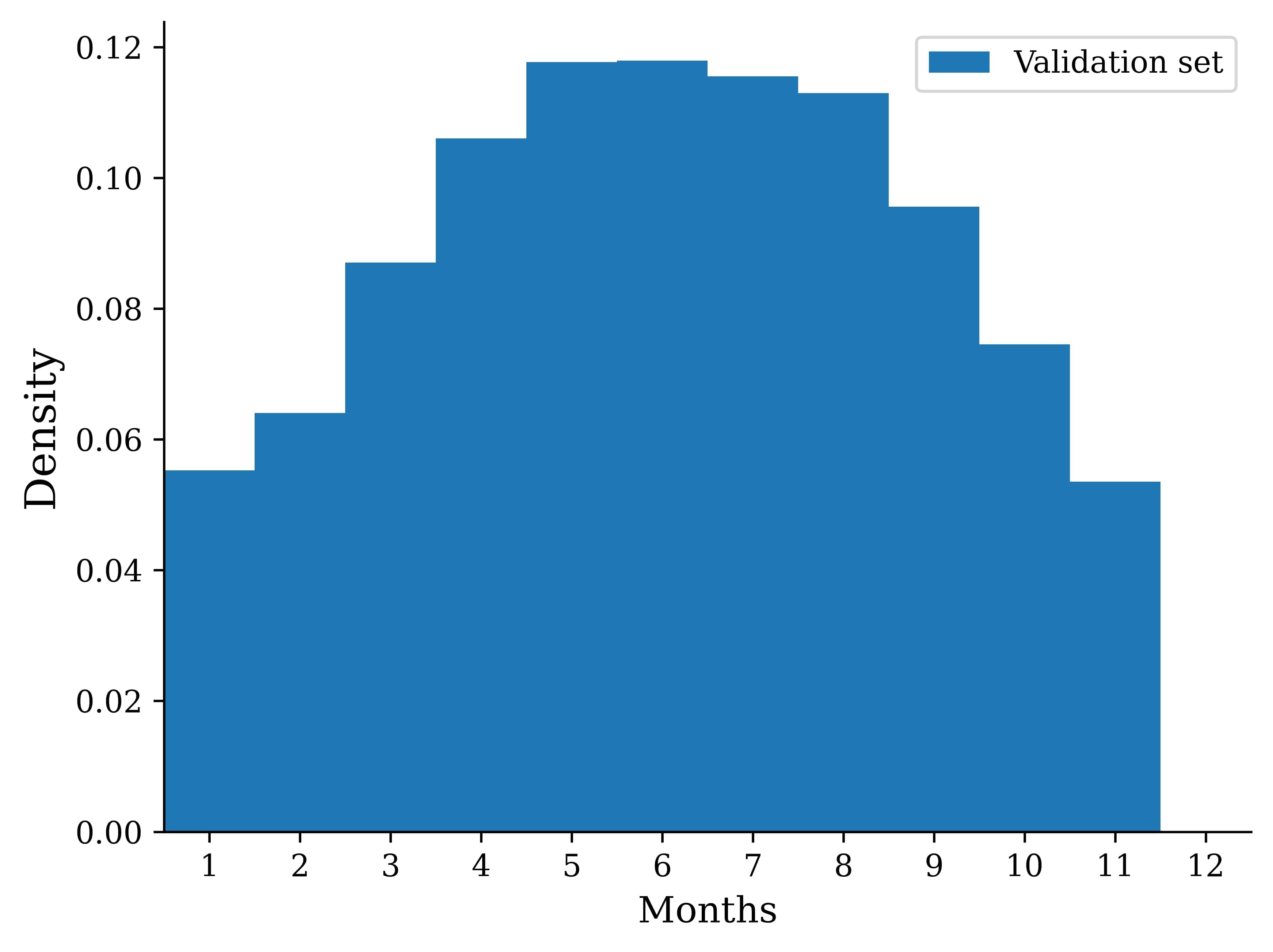}
    \label{fig:hist_months_val}
  \end{minipage}
  \begin{minipage}[b]{0.31\textwidth}
    \includegraphics[width=1\textwidth]{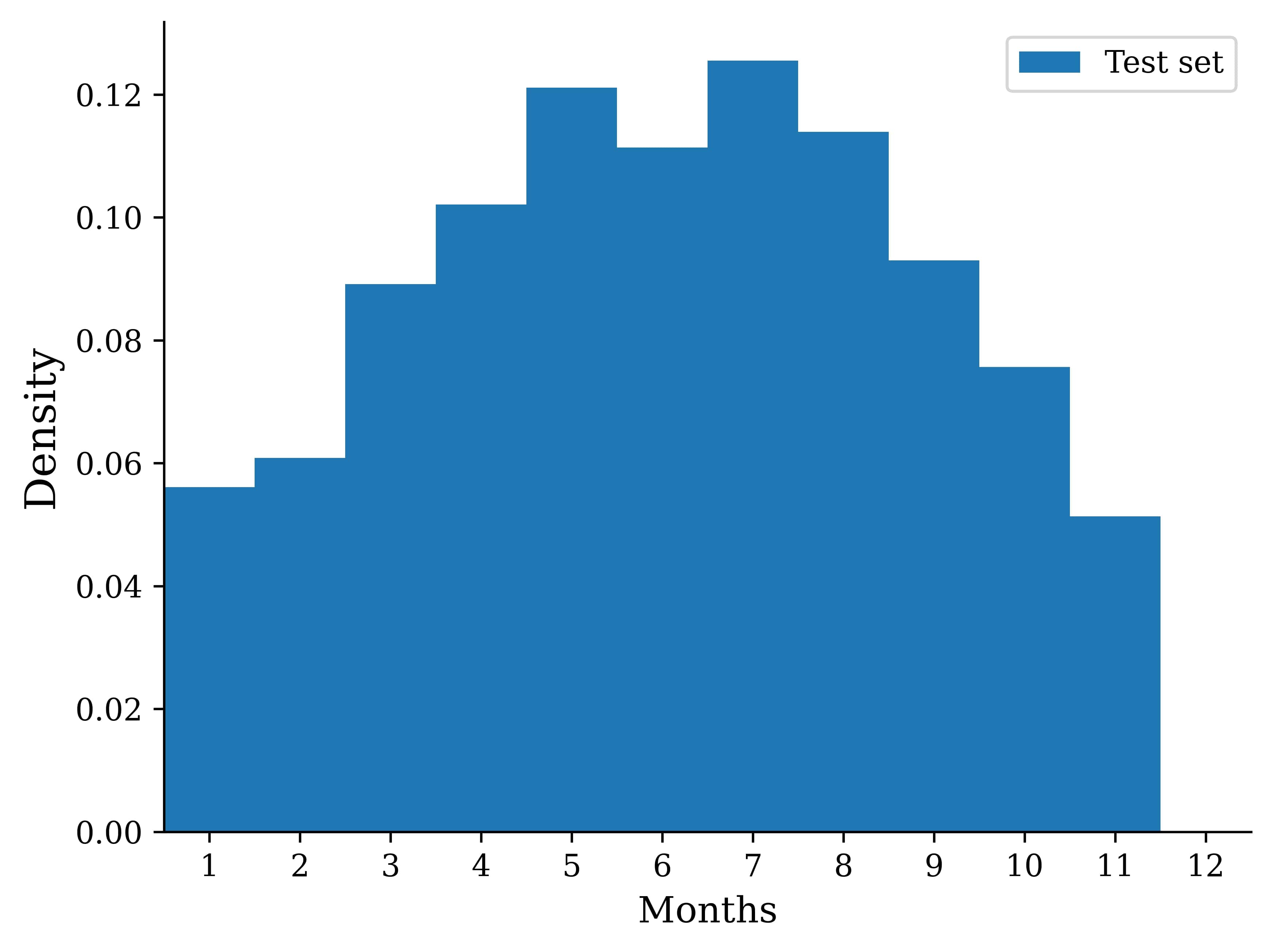}
    \label{fig:hist_months_test}
  \end{minipage}
  \vspace{-1\baselineskip}
\caption{Distribution of samples by months in the training, validation and test sets.}
\label{fig:dataset_distribution_month}
\end{figure}

\vspace{-0.5\baselineskip}

\begin{figure}[ht!]
\centering
\begin{minipage}[b]{0.31\textwidth}
    \includegraphics[width=1\textwidth]{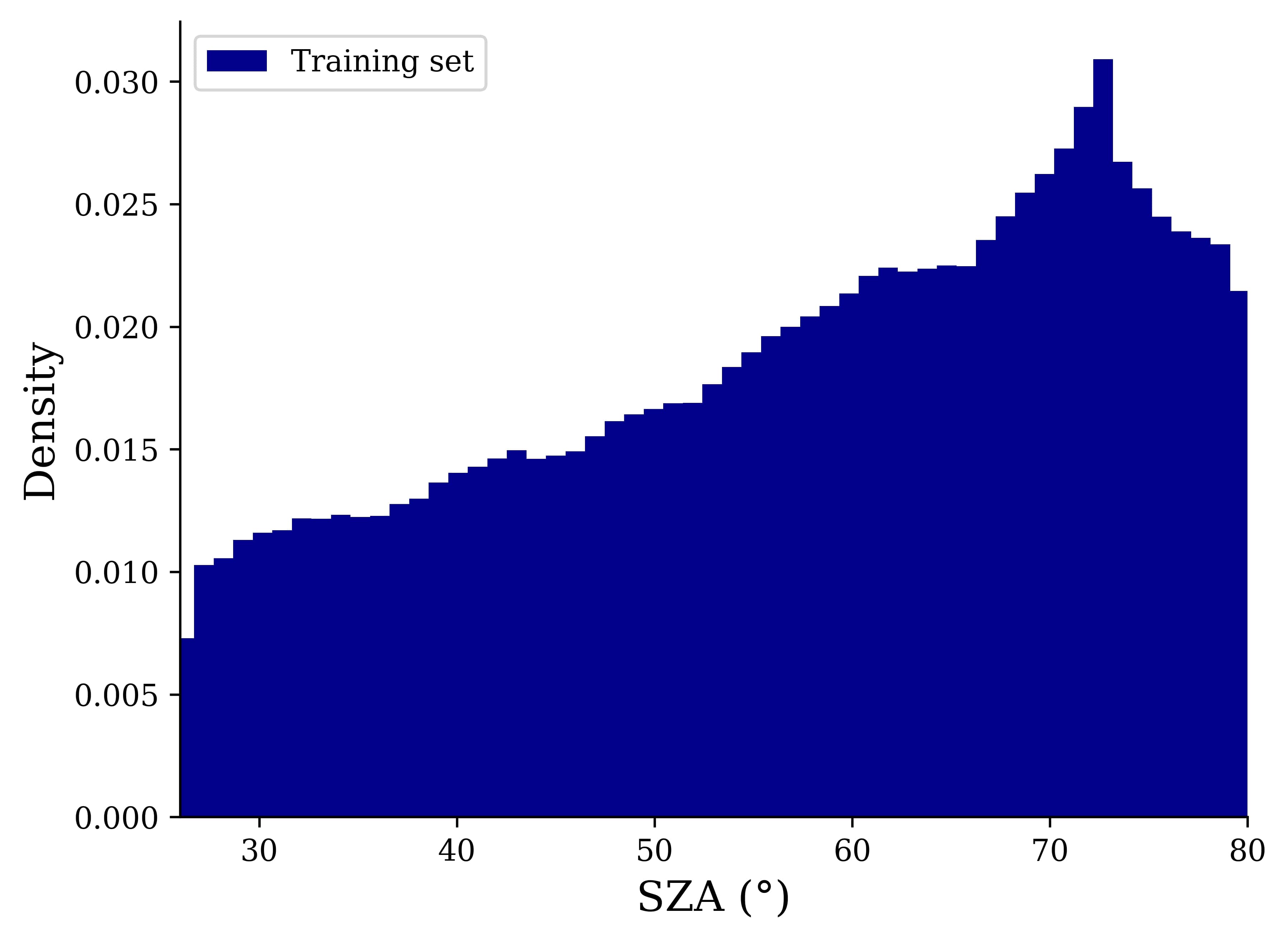}
    \label{fig:hist_sza_train}
  \end{minipage} 
  \begin{minipage}[b]{0.31\textwidth}
    \includegraphics[width=1\textwidth]{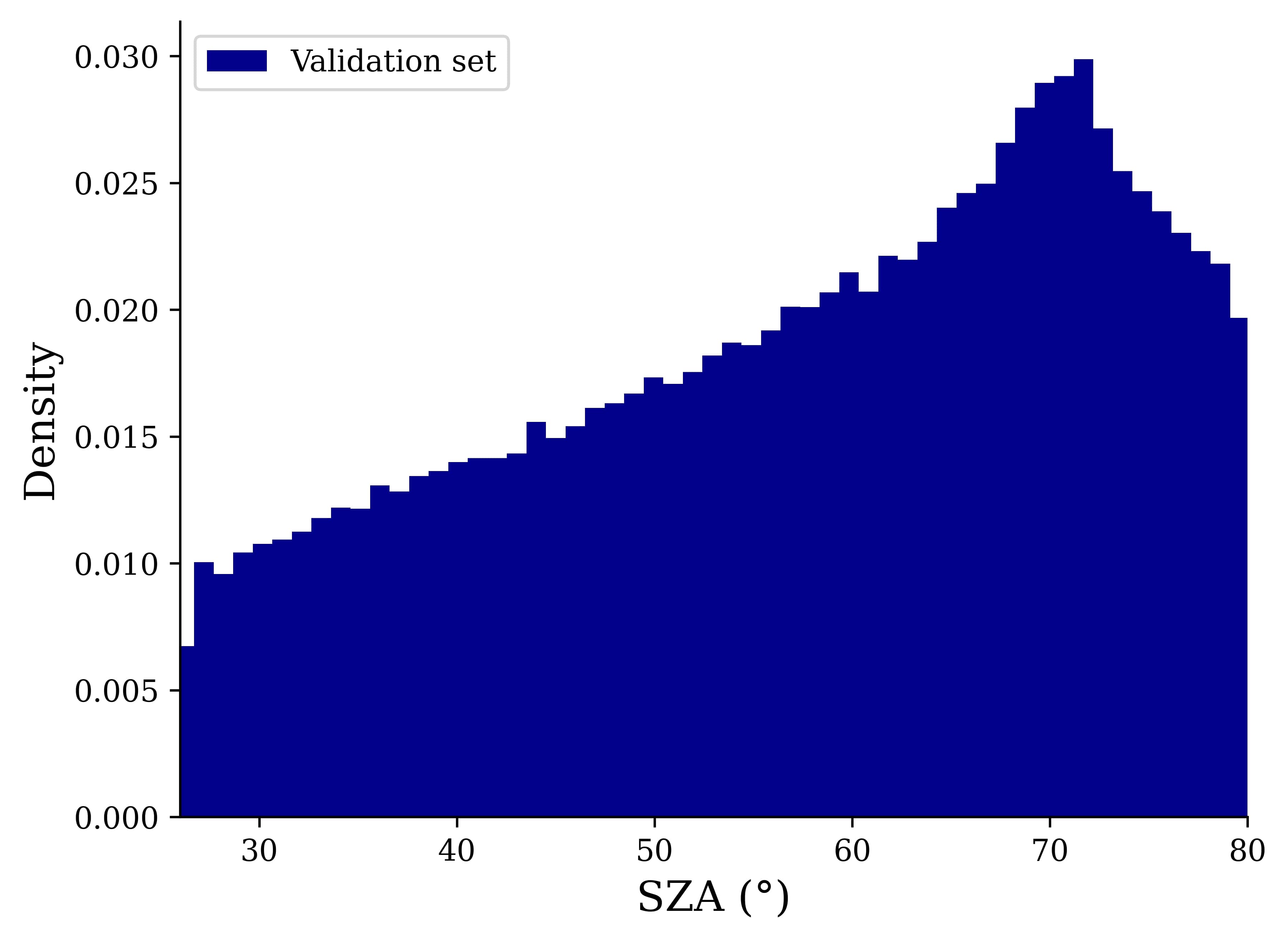}
    \label{fig:hist_sza_val}
  \end{minipage}
  \begin{minipage}[b]{0.31\textwidth}
    \includegraphics[width=1\textwidth]{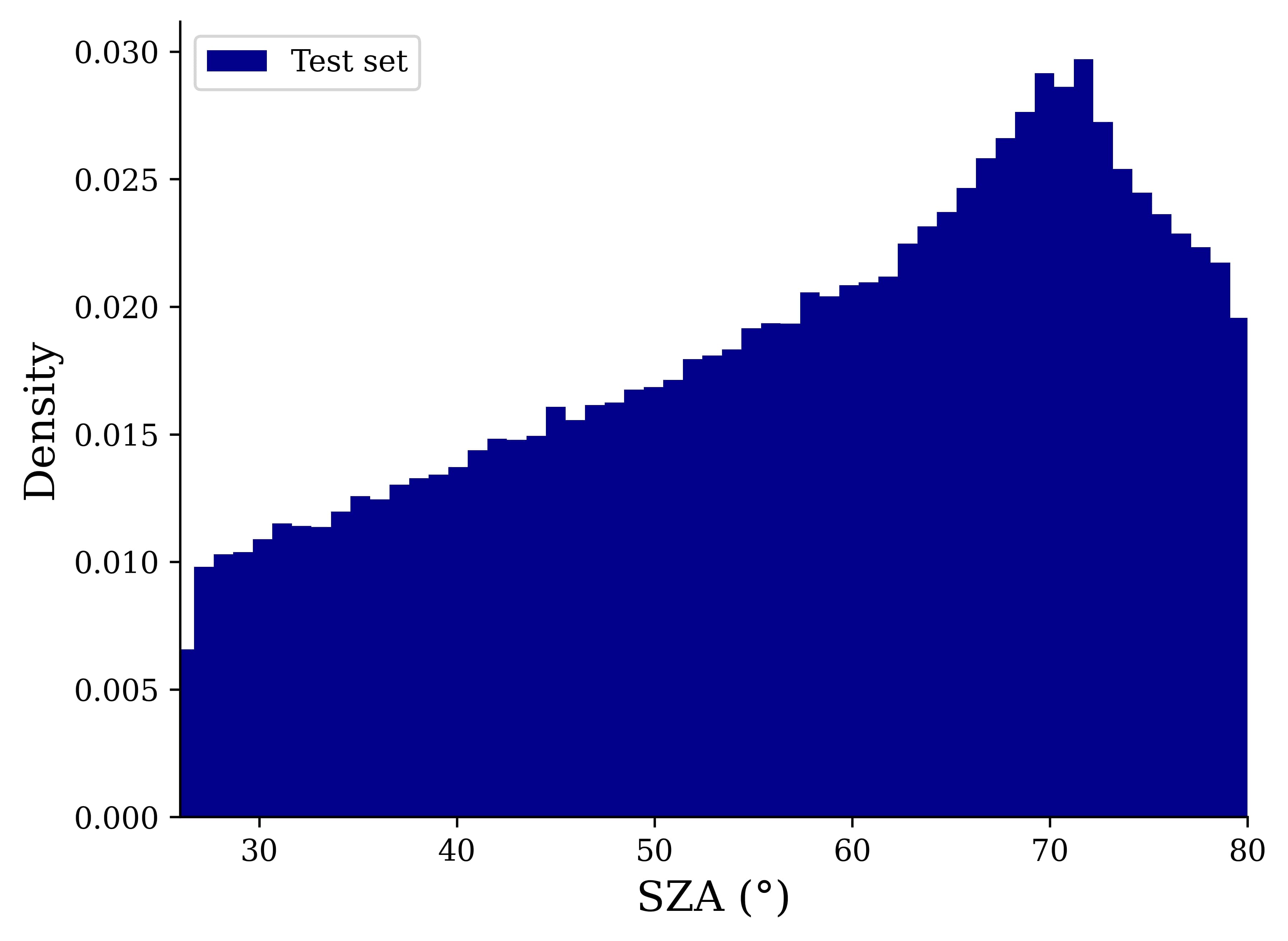}
    \label{fig:hist_sza_test}
  \end{minipage}
\vspace{-1\baselineskip}
\caption{Distribution of samples by Solar Zenith Angle in the training, validation and test sets.}
\label{fig:dataset_distribution_sza}
\end{figure}

\vspace{0.5\baselineskip}

\begin{figure}[ht!]
\centering
\begin{minipage}[b]{0.31\textwidth}
    \includegraphics[width=1\textwidth]{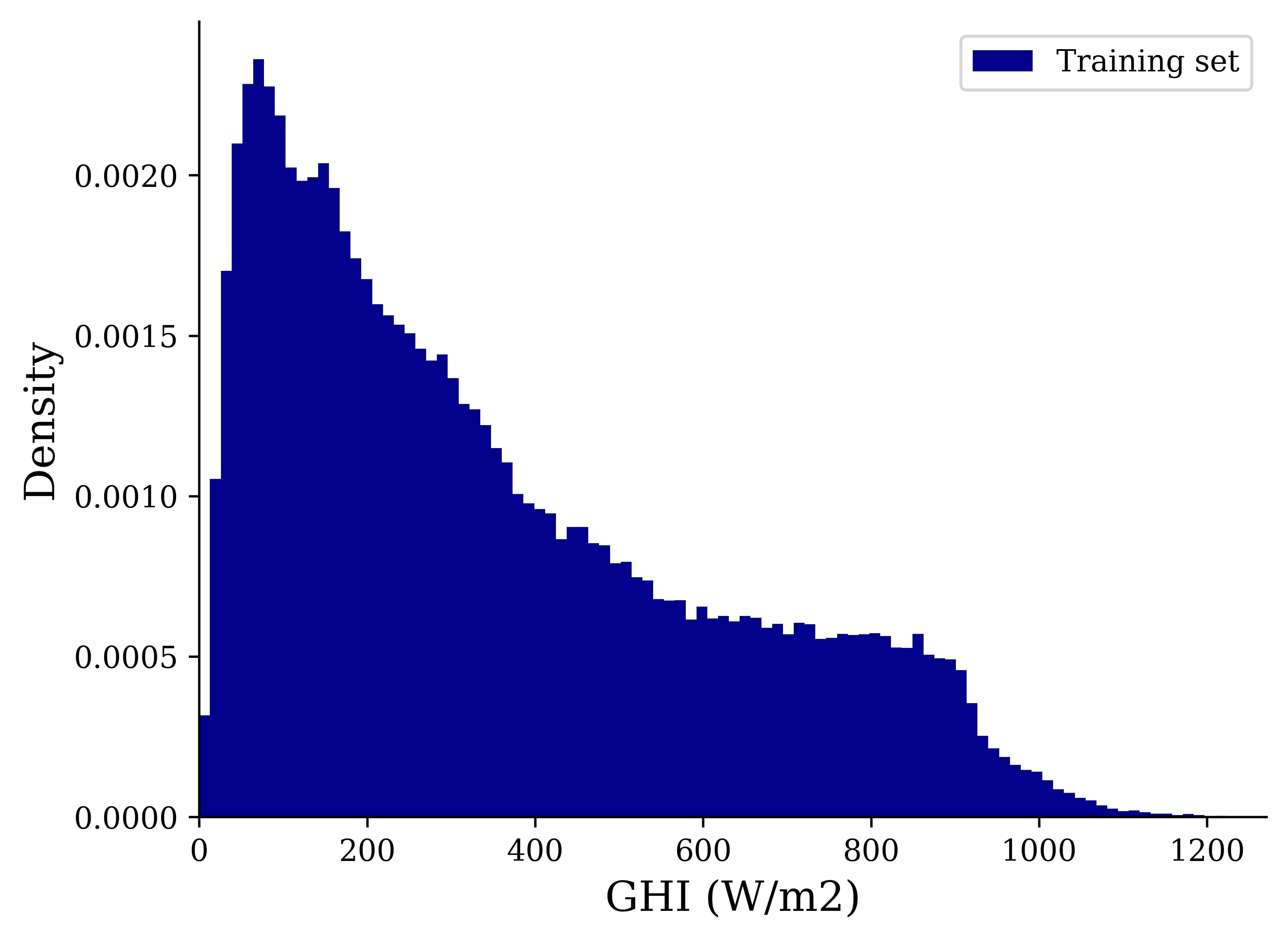}
    \label{fig:hist_bins_train}
  \end{minipage} 
  \begin{minipage}[b]{0.31\textwidth}
    \includegraphics[width=1\textwidth]{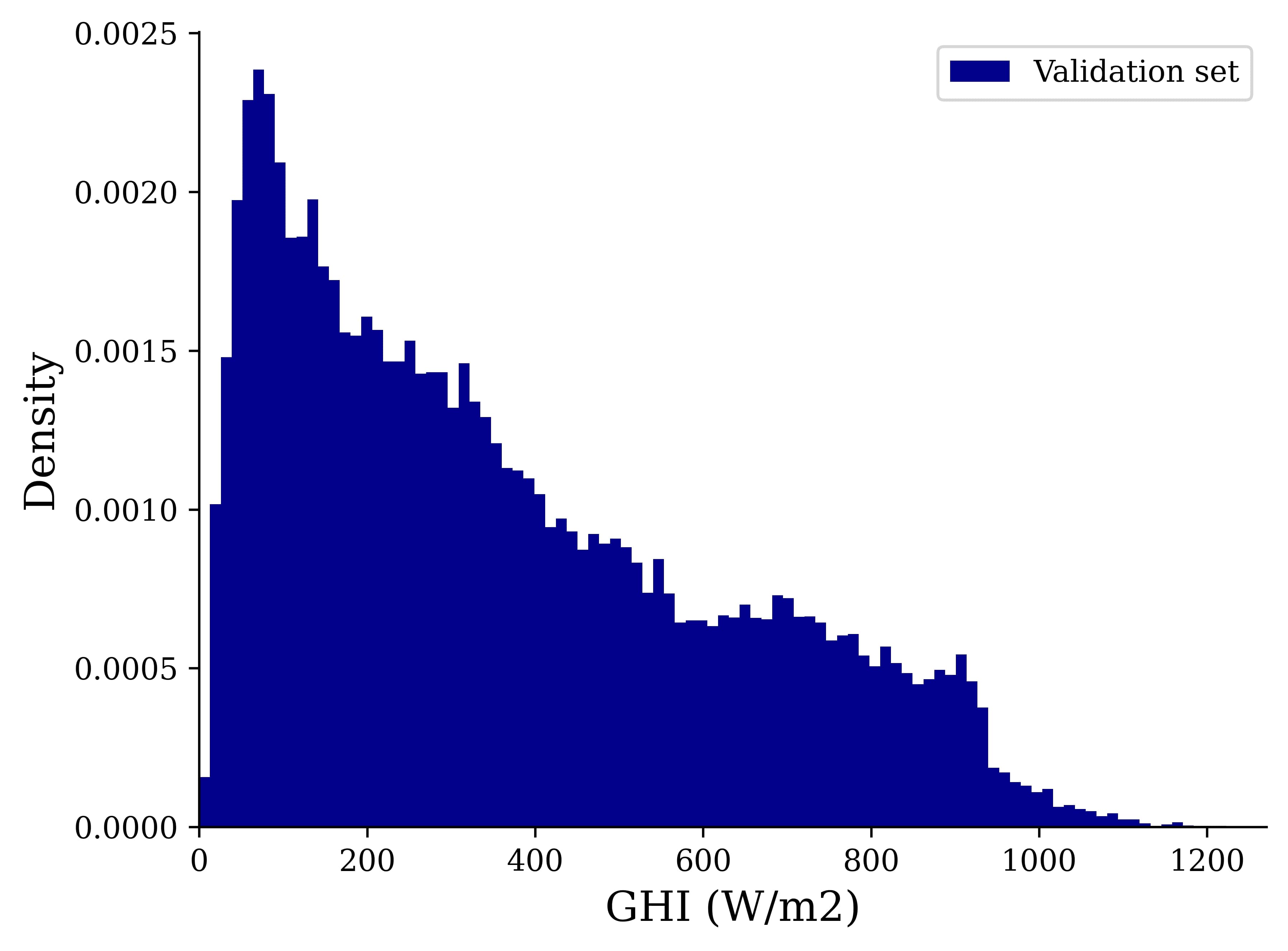}
    \label{fig:hist_bins_val}
  \end{minipage}
  \begin{minipage}[b]{0.31\textwidth}
    \includegraphics[width=1\textwidth]{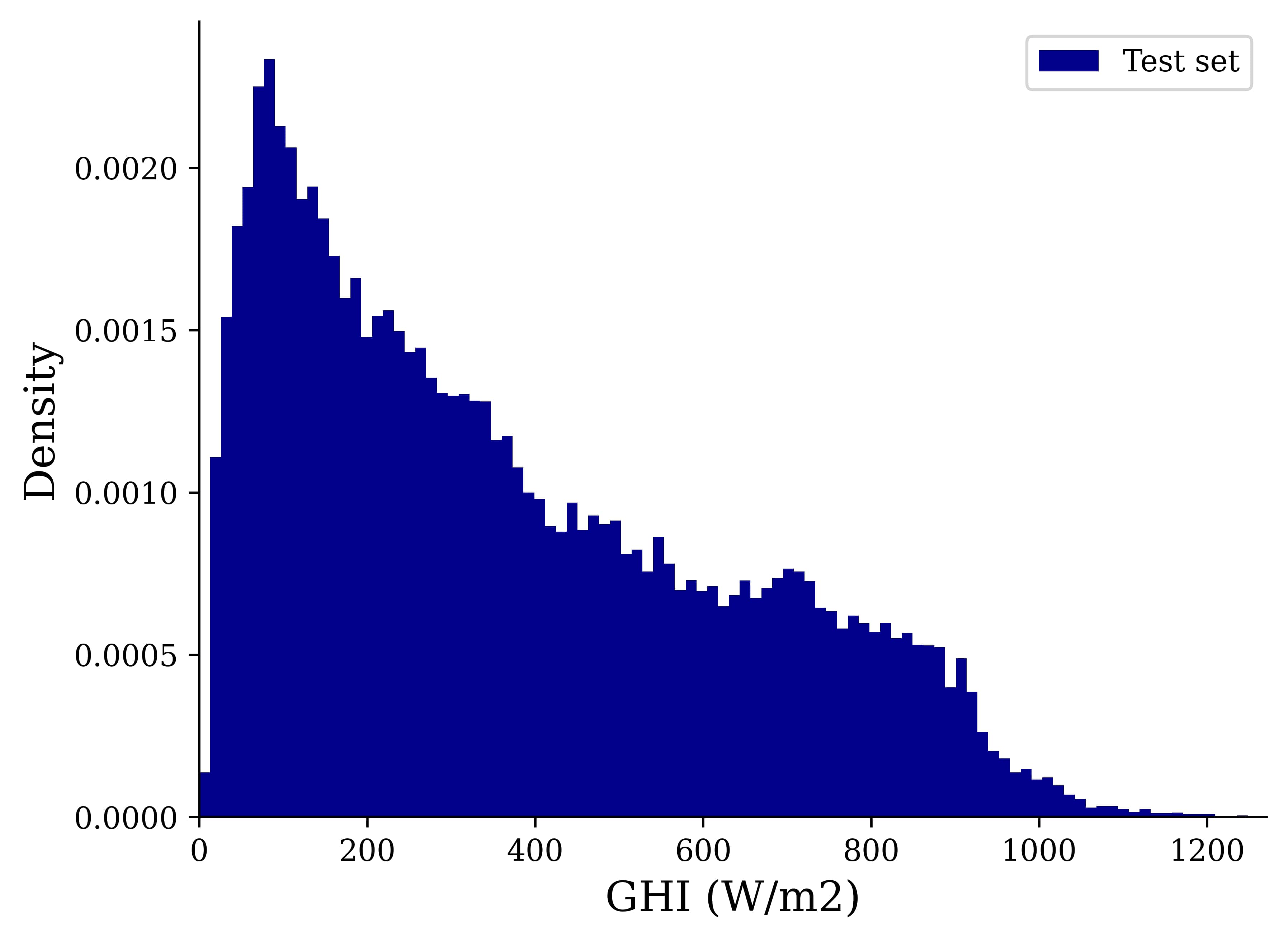}
    \label{fig:hist_bins_test}
  \end{minipage}
  \vspace{-1\baselineskip}
\caption{Distribution of samples by Global Horizontal Irradiance level in the training, validation and test sets.}
\label{fig:dataset_distribution_bins}
\end{figure}

\newpage

\section{Sun Tracking}
\label{section:sun_tracking}
\setcounter{figure}{0}
\setcounter{table}{0}

\begin{figure}[ht]
\centering    
\includegraphics[width=0.48\textwidth]{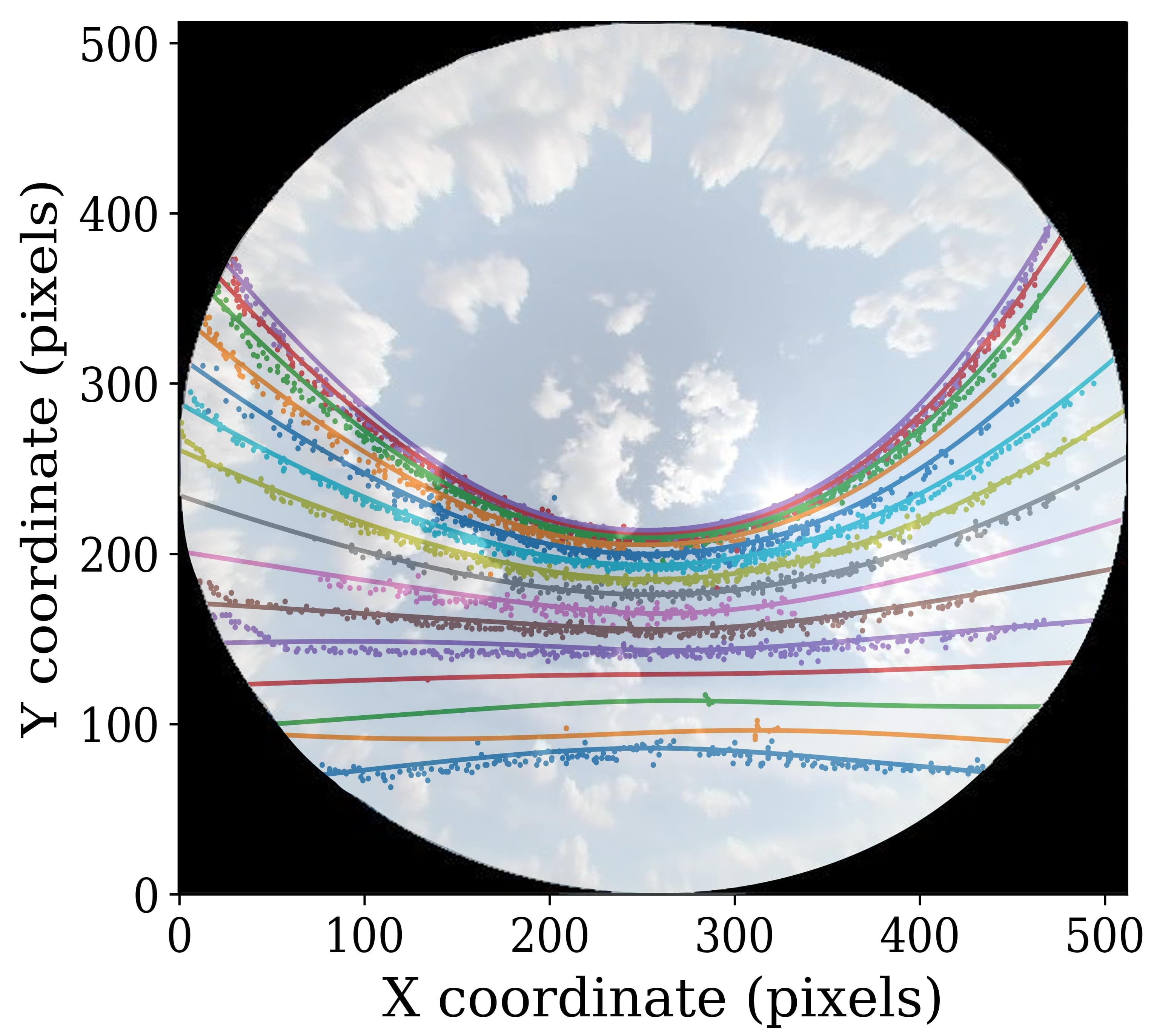}
\caption{Position of the sun in the sky for different days of the year (the 15 curves of the Figure correspond to 15 days sampled every 10 days from January to June 2017). Points correspond to observations of the sun and curves to sun trajectories over a day predicted by the algorithm from past observations. Taken from~\cite{palettaTemporallyConsistentImagebased2020}.}
\label{fig:sun_position_by_day_cleaned_15days_2017}
\end{figure}

The position of the sun in the image is obtained from an image-based sun tracking algorithm~\cite{palettaTemporallyConsistentImagebased2020} (3rd panel in Figure~\ref{fig:segmentation} and Figure~\ref{fig:sun_position_by_day_cleaned_15days_2017}). A decision rule, based on the presence of saturated pixels in the short exposure image, first classifies each image into two classes, `the sun position is visible' and `the sun position is not visible'. When the sun is visible, its position is extrapolated from saturated pixels~\cite{chusuntrackingImagingSystem2016, weiDesignSolarTracking2016}. To estimate the position of the sun when it is hidden, the overall trajectory of the sun is first modelled independently for each minute of the day given all observations.

\vspace{0.5\baselineskip}

Contrary to~\cite{palettaTemporallyConsistentImagebased2020}, for this step we use all observations (past and future) to fit a linear regression with a periodic basis (a basis composed of cosine and sine functions with a period of a year and half a year, respectively). Instead of removing outliers before estimating the trajectory of the sun over a day from the minute by minute estimates, we minimise an L1 loss function, which is less affected by the largest errors. 

\vspace{0.5\baselineskip}

Finally, a regularised polynomial regression provides a minute-by-minute smooth trajectory from estimates given by the first periodic regression. The Mean Absolute Error (MAE) between the final estimates resulting from this method and the visible sun observations over the entire dataset reached 0.7\% of the image's width compared to 0.9\% with the original method~\cite{palettaTemporallyConsistentImagebased2020}.

\section{Architecture \& Training}
\label{section:architecture}
\setcounter{figure}{0}
\setcounter{table}{0}

ECLIPSE's architecture is made up of 4 modules comprising a total number of 7.70M parameters:
\vspace{0.5\baselineskip}

- Spatial encoder: 0.68M parameters

- Temporal encoder: 0.46M parameters

- Future state prediction module: 4.91M parameters

- Decoders: 1.64M parameters

\vspace{0.5\baselineskip}
ECLIPSE was trained on a Nvidia Tesla V100 for about 150,000 steps (15h) with a batch size of 10. The optimiser is Adam with a learning rate of $2.5   \times 10^{-4}$. In this work, ECLIPSE is fed with the same past context as the other benchmarked models, i.e. 8-min (5 frames), to predict the next 10-min (5 frames). The python library Pytorch was used to implement the model.

\newpage

\section{Auxiliary data and temporal lag}
\label{section:temporal_lag}
\setcounter{figure}{0}
\setcounter{table}{0}

As well as sky images, a range of auxiliary data can be integrated into solar irradiance modelling. The most common variables are past target measurements (irradiance, PV output, etc.) and the position of the sun (solar zenith / azimuthal angles and their transformations: sine, cosine, etc.). We compare in Table~\ref{tab:results_temporal_lag_fs} the effect of each input signal on the forecasting performance when integrated to the predictions through a specific parallel encoder for auxiliary data (densely connected layers). The ConvLSTM model is fed both video streams and auxiliary data in parallel networks merged into one as in ~\cite{palettaBenchmarkingDeepLearning2021c}.
\vspace{-0.5\baselineskip}

\begin{table*}[ht!]
\begin{center}
\begin{tabular}{lcccccccc}
\hline
\noalign{\vskip 1mm}
 & & \multicolumn{3}{c}{Forecast Skill $\uparrow$ [\%]} && \multicolumn{3}{c}{TDI $\downarrow$ [\%] (Advance / Late)}\\
 Forecast Horizon & $\mid$ & 2-min & 6-min & 10-min & $\mid$ & 2-min & 6-min & 10-min \\
\hline\hline
\noalign{\vskip 1mm}
No Auxiliary data && -3.4\% & 16.8\% & 18.4\% && 7.8\% (3.8 / 4.0) & 9.5\% (4.0 / 5.5) & 10.7\% (3.9 / 6.8) \\
Sun position && -2.4\% & 19.8\% & 21.6\% && 7.5\% (3.8 / 3.7) & 8.9\% (4.1 / 4.8) & 10.0\% (4.2 / \textbf{5.8}) \\
Irradiance && \textbf{12.9\%} & 22.2\% & 22.4\% && \textbf{3.1\%} \textbf{(0.9 / 2.2)} & \textbf{6.4\%} (\textbf{2.0} / 4.4) & 8.7\% (\textbf{2.7} / 6.0) \\
Irradiance + Sun position && 12.3\% & \textbf{23.0\%} &  \textbf{23.5\%} && 3.7\% (1.4 / 2.3) & \textbf{6.4\%} (2.1 / \textbf{4.2}) & \textbf{8.5\%} \textbf{(2.7 / 5.8)} \\
\noalign{\vskip 1mm}
\hline
\end{tabular}
\end{center}
\vspace{-1\baselineskip}
\caption{Quantitative performance of the ConvLSTM model for different types of auxiliary data: none, past irradiance, sun position, past irradiance and sun position. Results are averaged over 5 trainings.}
\label{tab:results_temporal_lag_fs}
\end{table*}

Regarding the FS metrics, adding both auxiliary data (past irradiance measurements and the sun position) results in the largest gains on the 6 and 10-min ahead predictions. Interestingly, providing past irradiance measurements significantly improves short-term forecasting from -3.4 to 12.9\%. This indicates that the model struggles to correlate an image with its corresponding irradiance level~\cite{sunSolarPVOutput2018}. Given that a majority of the input solar flux directly originates from the sun (direct normal irradiance), the circumsolar area is key to approximate the contemporaneous irradiance level. However, this region is relatively small (a few percent of the original image) and often saturated when the sun is visible. As a consequence the information is hard to retrieve. Learning from a close-up on the circumsolar area~\cite{palettaSPIN2021} or from images taken with different exposition times or wavelengths might benefit image to irradiance translation.

\vspace{0.3\baselineskip}

Furthermore, learning from past irradiance measurements tends to significantly decrease temporal distortion, especially early predictions ($\text{TDI}_{adv}$). Models using the current irradiance are less likely to predict a trend too early as it would not match the current irradiance level. 
\vspace{-1.0\baselineskip}

\section{Predictions based on past irradiance measurements}
\label{section:past_irradiance}
\setcounter{figure}{0}
\setcounter{table}{0}

\vspace{-0.5\baselineskip}

\ref{tab:results_past_irradiance} highlights the performance of the four benchmarked models with this new approach (predicting irradiance change from RGBI inputs). For models learning from both video and time series (with an auxiliary data encoder), inputting the past irradiance level closer to the end of the network through densely connected layers instead of adding an additional irradiance channel, appears to result in slightly improved performance: the FS of the ConvLSTM increases on all horizons, especially on the 2-min ahead forecast, from -3.4\% without the information, 12.1\% with the irradiance channel (and irradiance change prediction) to 12.9\% when the information is integrated closer to the prediction head through densely connected layers merged with the video decoder output. See~\cite{palettaBenchmarkingDeepLearning2021c} for more details on the ConvLSTM architecture used in this study.

\vspace{-0.5\baselineskip}

\begin{table*}[ht!]
\begin{center}
\begin{tabular}{lcccccccc}
\hline
\noalign{\vskip 1mm}
 & & \multicolumn{3}{c}{RMSE $\downarrow$ [W/$\text{m}^2$] (Forecast Skill $\uparrow$ [\%])} & & \multicolumn{3}{c}{TDI $\downarrow$ [\%] (Advance / Late)}\\
 \noalign{\vskip 1mm}
 Forecast Horizon & $\mid$ & 2-min & 6-min & 10-min & $\mid$ & 2-min & 6-min & 10-min \\
\hline\hline
\noalign{\vskip 1mm}
Smart Pers. && 93.3 (0\%) & 129.0 (0\%) & 143.6 (0\%) && 1.9 (0.0/1.9) & 5.2 (0.0/5.2) & 8.4 (0.0/8.4) \\
ConvLSTM~\cite{palettaBenchmarkingDeepLearning2021c} && 82.0 (12.1\%) & 101.8 (21.1\%) & 111.7 (22.2\%) && 4.3 (1.3/2.9) & 7.6 (2.2/5.3) & 10.2 (3.2/7.0) \\
TimeSFormer~\cite{bertasiusSpaceTimeAttentionAll2021} && 80.2 (14.0\%) & 100.2 (22.3\%) & 109.6 (23.7\%) && \textbf{4.1} (\textbf{1.3}/2.8) & \textbf{7.3} (2.6/\textbf{4.7}) & 10.8 (3.8/7.0) \\
PhyDNet~\cite{leguenDeepPhysicalModel2020a} && 77.3 (17.1\%) & 98.4 (23.7\%) & 109.4 (23.8\%) && 5.3 (2.0/3.3) & 8.6 (\textbf{2.5}/6.1) & 14.4 (5.2/9.1) \\
ECLIPSE && \textbf{76.0 (18.5\%)} & \textbf{95.3 (26.1\%)} & \textbf{105.9 (26.3\%)} && 4.2 (1.6/\textbf{2.6}) & 7.9 (2.9/5.0) & \textbf{9.5 (3.1/6.3)} \\
\noalign{\vskip 1mm}
\hline
\end{tabular}
\end{center}
\vspace{-1.0\baselineskip}
\caption{In addition to the three RGB channels, the contemporaneous irradiance level is given to the models through a fourth channel (RGBI). Models are trained to forecast the future irradiance change instead of the absolute irradiance.}
\label{tab:results_past_irradiance}
\end{table*}

\begin{figure}[ht!]
\centering    

\includegraphics[width=0.55\textwidth]{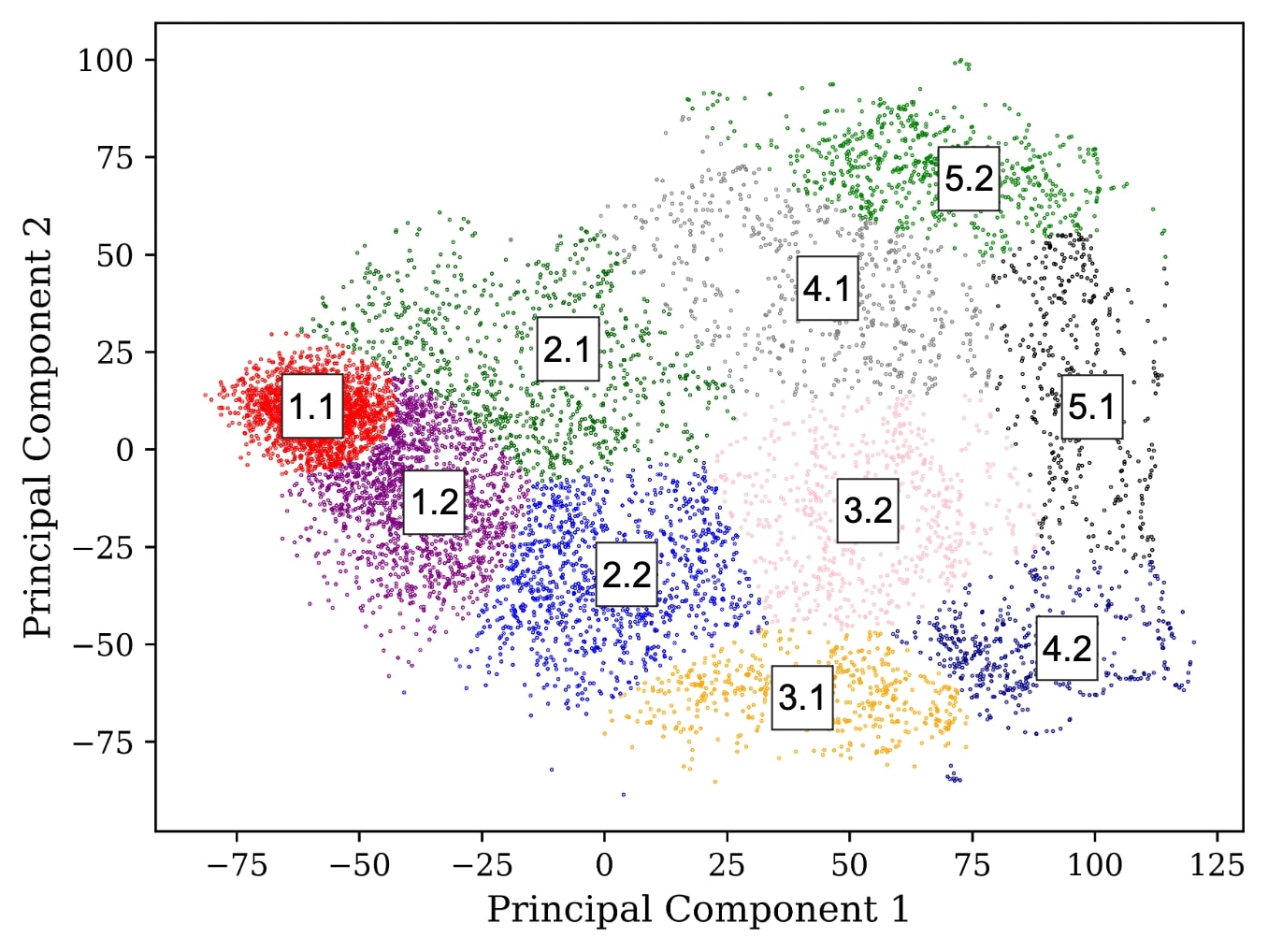}
\vspace{0.4\baselineskip}

\begin{minipage}[b]{0.15\textwidth}
    \includegraphics[width=1\textwidth]{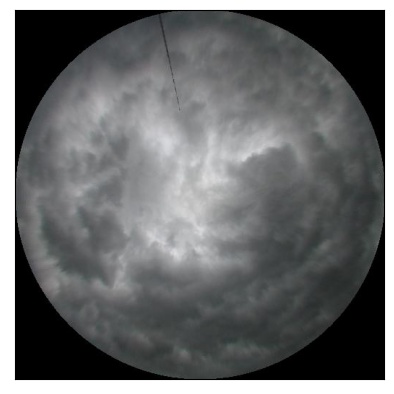}
\end{minipage} 
\begin{minipage}[b]{0.15\textwidth}
    \includegraphics[width=1\textwidth]{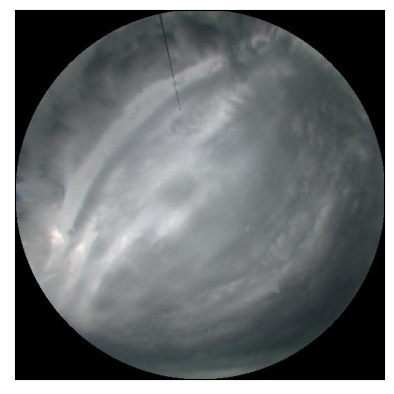}
\end{minipage}
\begin{minipage}[b]{0.15\textwidth}
    \includegraphics[width=1\textwidth]{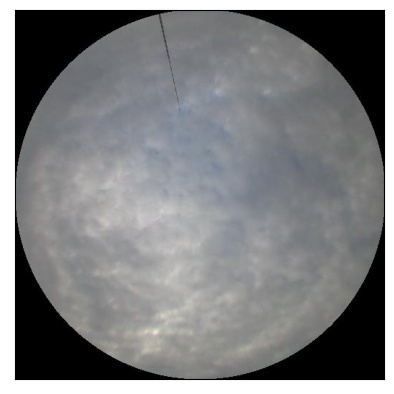}
\end{minipage}
\hspace{0.8cm}
\begin{minipage}[b]{0.15\textwidth}
    \includegraphics[width=1\textwidth]{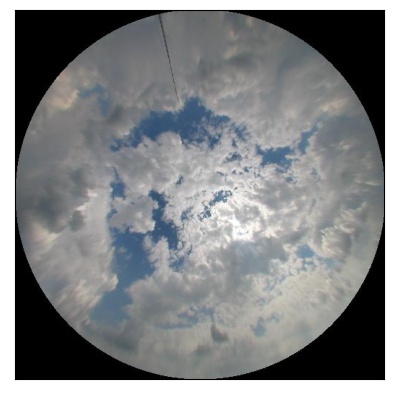}
\end{minipage}
\begin{minipage}[b]{0.15\textwidth}
    \includegraphics[width=1\textwidth]{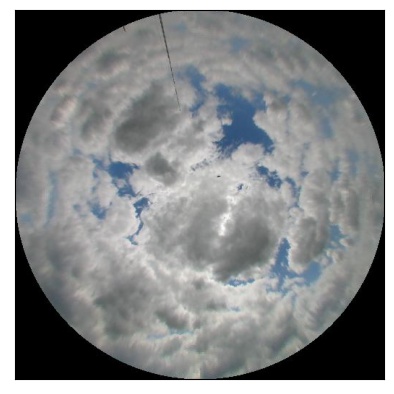}
\end{minipage} 
\begin{minipage}[b]{0.15\textwidth}
    \includegraphics[width=1\textwidth]{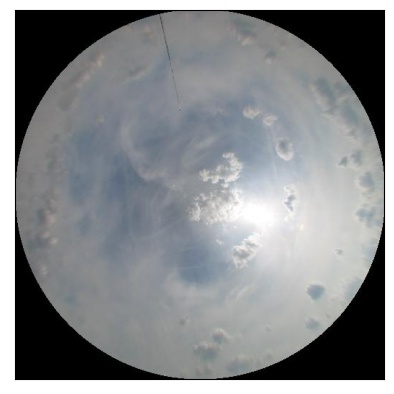}
\end{minipage}

\begin{minipage}[b]{0.15\textwidth}
    \includegraphics[width=1\textwidth]{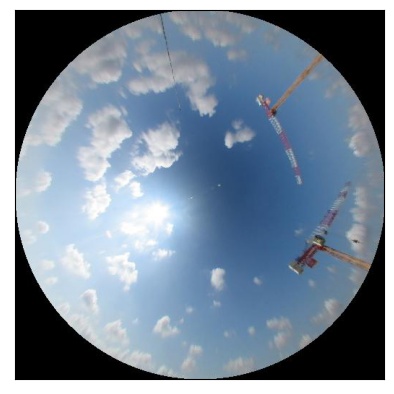}
\end{minipage} 
\begin{minipage}[b]{0.15\textwidth}
    \includegraphics[width=1\textwidth]{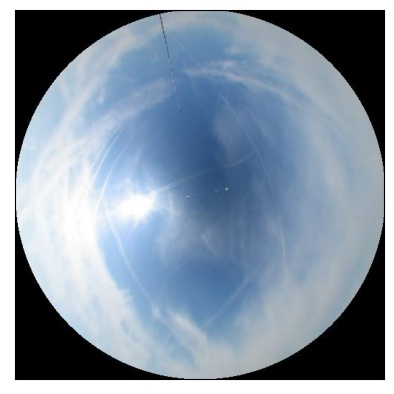}
\end{minipage}
\begin{minipage}[b]{0.15\textwidth}
    \includegraphics[width=1\textwidth]{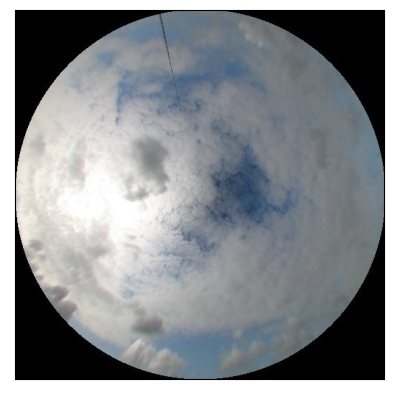}
\end{minipage}
\hspace{0.8cm}
\begin{minipage}[b]{0.15\textwidth}
    \includegraphics[width=1\textwidth]{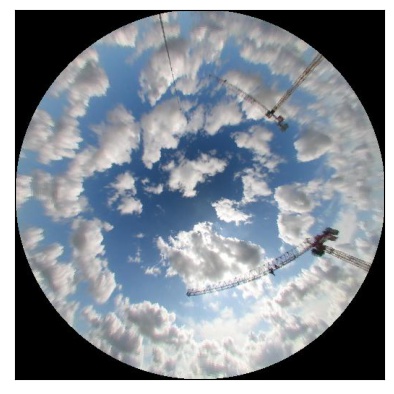}
\end{minipage}
\begin{minipage}[b]{0.15\textwidth}
    \includegraphics[width=1\textwidth]{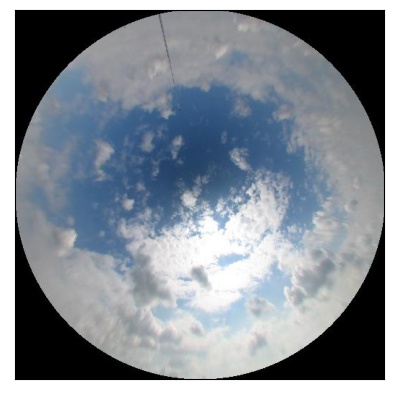}
\end{minipage} 
\begin{minipage}[b]{0.15\textwidth}
    \includegraphics[width=1\textwidth]{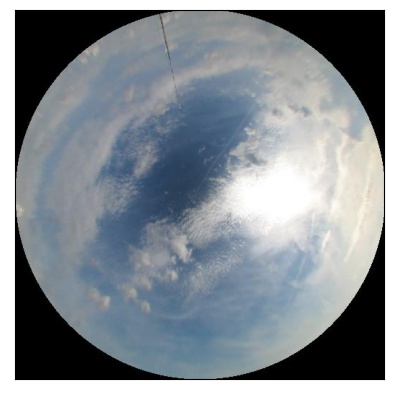}
\end{minipage}

\begin{minipage}[b]{0.15\textwidth}
    \includegraphics[width=1\textwidth]{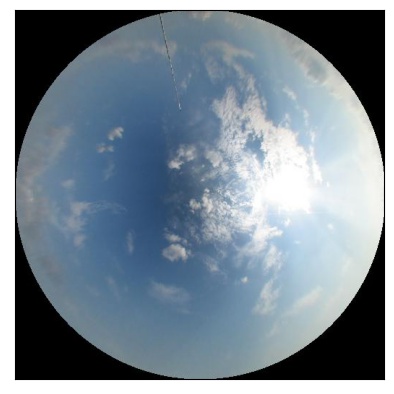}
\end{minipage} 
\begin{minipage}[b]{0.15\textwidth}
    \includegraphics[width=1\textwidth]{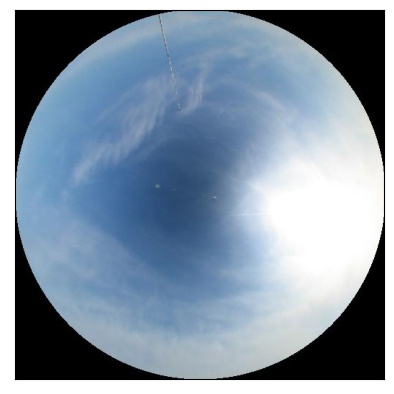}
\end{minipage}
\begin{minipage}[b]{0.15\textwidth}
    \includegraphics[width=1\textwidth]{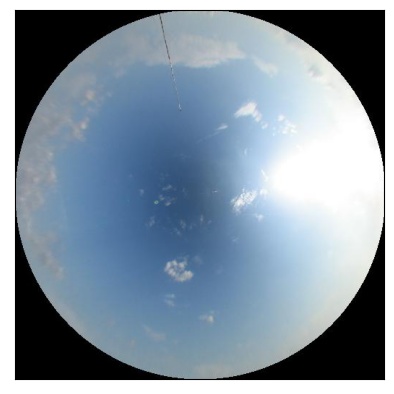}
\end{minipage}
\hspace{0.8cm}
\begin{minipage}[b]{0.15\textwidth}
    \includegraphics[width=1\textwidth]{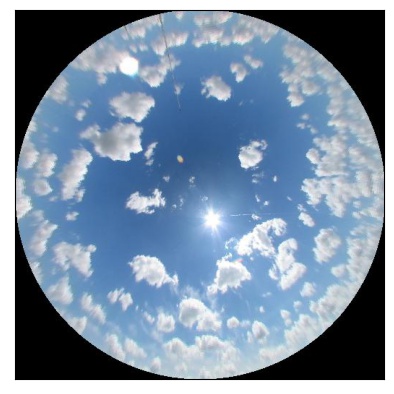}
\end{minipage}
\begin{minipage}[b]{0.15\textwidth}
    \includegraphics[width=1\textwidth]{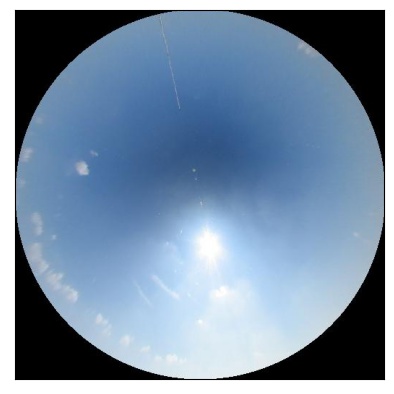}
\end{minipage} 
\begin{minipage}[b]{0.15\textwidth}
    \includegraphics[width=1\textwidth]{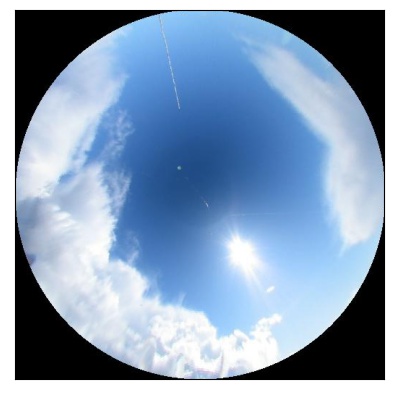}
\end{minipage}

\begin{minipage}[b]{0.15\textwidth}
    \includegraphics[width=1\textwidth]{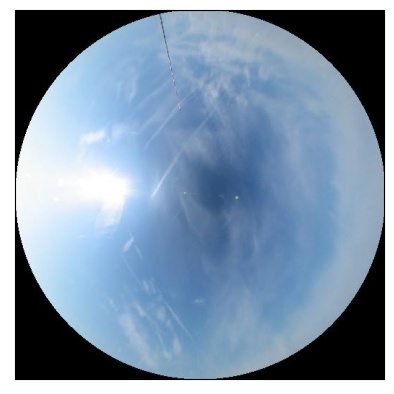}
\end{minipage} 
\begin{minipage}[b]{0.15\textwidth}
    \includegraphics[width=1\textwidth]{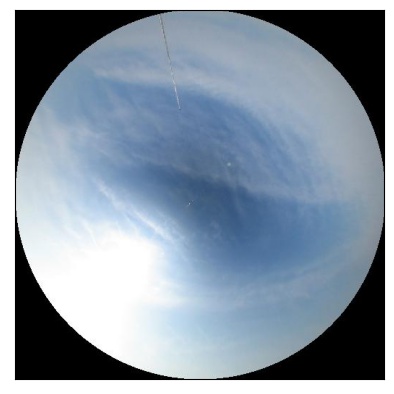}
\end{minipage}
\begin{minipage}[b]{0.15\textwidth}
    \includegraphics[width=1\textwidth]{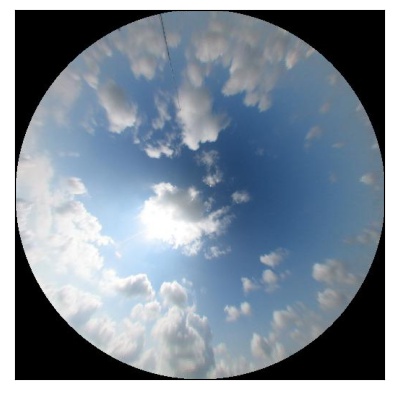}
\end{minipage}
\hspace{0.8cm}
\begin{minipage}[b]{0.15\textwidth}
    \includegraphics[width=1\textwidth]{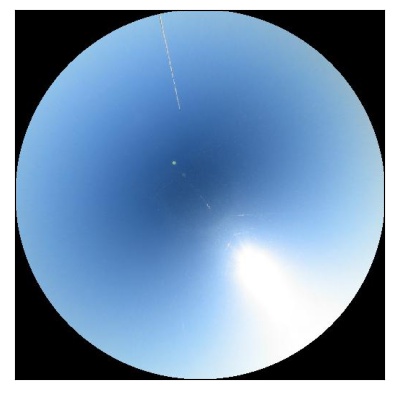}
\end{minipage}
\begin{minipage}[b]{0.15\textwidth}
    \includegraphics[width=1\textwidth]{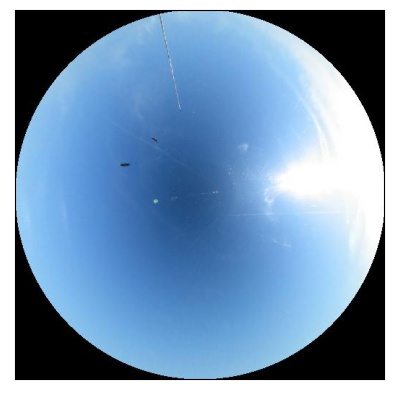}
\end{minipage} 
\begin{minipage}[b]{0.15\textwidth}
    \includegraphics[width=1\textwidth]{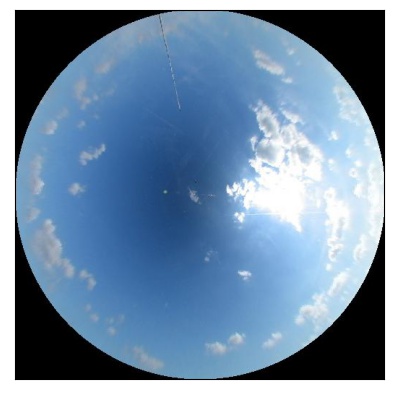}
\end{minipage}

\begin{minipage}[b]{0.15\textwidth}
    \includegraphics[width=1\textwidth]{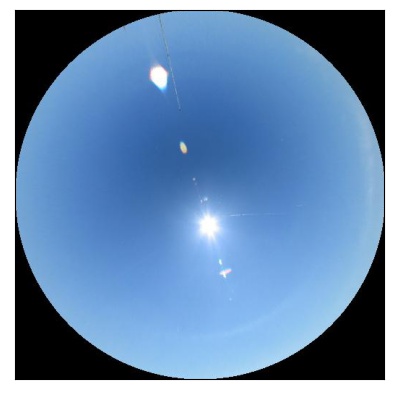}
\end{minipage} 
\begin{minipage}[b]{0.15\textwidth}
    \includegraphics[width=1\textwidth]{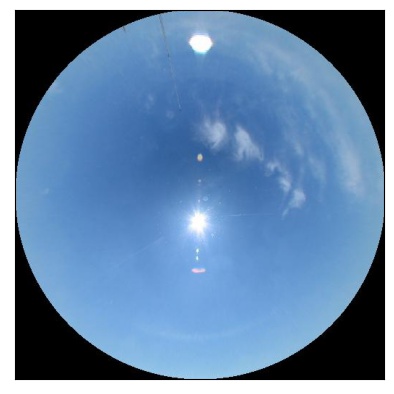}
\end{minipage}
\begin{minipage}[b]{0.15\textwidth}
    \includegraphics[width=1\textwidth]{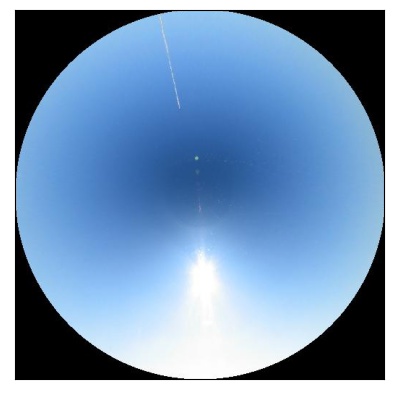}
\end{minipage}
\hspace{0.8cm}
\begin{minipage}[b]{0.15\textwidth}
    \includegraphics[width=1\textwidth]{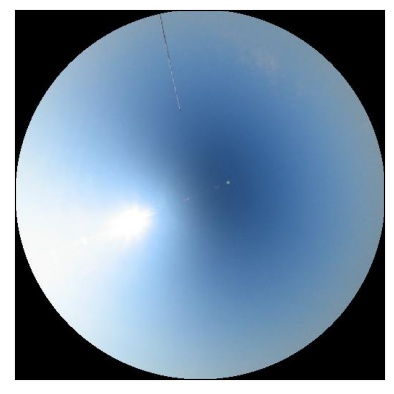}
\end{minipage}
\begin{minipage}[b]{0.15\textwidth}
    \includegraphics[width=1\textwidth]{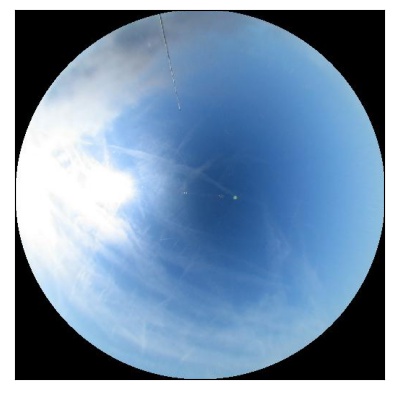}
\end{minipage} 
\begin{minipage}[b]{0.15\textwidth}
    \includegraphics[width=1\textwidth]{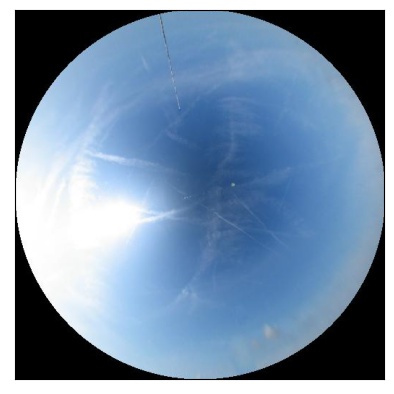}
\end{minipage}

\vspace{0.0\baselineskip}

\caption{Distribution of the first two components of the temporal representation (state $z_t$ in Figure~\ref{fig:model_architecture}) resulting from a principal component analysis on 9000 test samples followed by a clustering using a Gaussian Mixture model into 10 classes highlighted by different colours in the graph above. As illustrated in Figure~\ref{fig:pca_components_examples}, the first two components correspond to the extend of the cloud coverage (PC1) and the horizontal position of the sun (PC2). To illustrate this classification, three samples corresponding to each cluster are presented below (First column: 1.1, 2.1, 3.1, 4.1, 5.1, Second column: 1.2, 2.2, 3.2, 4.2, 5.2).}
\label{fig:pca_12_clusters_components_examples}
\end{figure}

\end{document}